\documentclass[a4,11pt]{article}
\usepackage[margin=3cm]{geometry}
\usepackage{natbib}
\usepackage{supertabular}
\usepackage{graphicx}
\usepackage[figuresright]{rotating}
\usepackage{epsfig}
\newcommand{\1}{{\rm 1}\kern-0.28em{\rm I}}
\renewcommand{\P}{{\rm I}\kern-0.22em{\rm P}}
\newcommand{\E}{{\rm I}\kern-0.22em{\rm E}}

\newcommand{\bx}{{\bf x}}
\newcommand{\bX}{{\bf X}}
\newcommand{\bbX}{{\bf {\mathcal X}}}

\newcommand{\bZ}{{\bf Z}}
\newcommand{\R}{{\rm I}\kern-0.20em{\rm R}}
\newcommand{\BEQ}{\begin{equation}}
\newcommand{\EEQ}{\end{equation}}
\newcommand{\BA}{\begin{array}{ll}}
\newcommand{\EA}{\end{array}}
\newcommand{\BP}{\begin{pmatrix}}
\newcommand{\EP}{\end{pmatrix}}

\begin{document}

\title{{\bf An empirical comparative study of approximate methods for binary graphical
models; application to the search of associations among causes of
death in French death certificates.}}

\author{Vivian Viallon , Onureena Banerjee\\
       Gr\'egoire Rey , Eric Jougla  \\
       Jo\"el Coste }

\date{}

\maketitle

\begin{abstract}
Looking for associations among multiple variables is a topical
issue in statistics due to the increasing amount of data
encountered in biology, medicine and many other domains involving
statistical applications. Graphical models have recently gained
popularity for this purpose in the statistical literature.
Following the ideas of the LASSO procedure designed for the linear
regression framework, recent developments dealing with graphical
model selection have been based on $\ell_1$-penalization. In the
binary case, however, exact inference is generally very slow or
even intractable because of the form of the so-called
log-partition function. Various approximate methods have recently
been proposed in the literature and the main objective of this
paper is to compare them. Through an extensive simulation study,
we show that a simple modification of a method relying on a
Gaussian approximation achieves good performance and is very fast.
We present a real application in which we search for associations
among causes of death recorded on French death certificates.
\end{abstract}


\section{Introduction}

In biology, medicine, and many other domains of statistical
application, researchers are increasingly faced with problems
involving numerous variables and a natural problem is that of
studying their relationships. Standard examples are the
construction of social or communication networks and systems
biology. When the underlying variables are binary (which is the
focus of this paper), a classical way for studying their
relationships is to use \emph{Poisson log-linear models} for
multiway \emph{contingency tables}
\citep{Agresti}\citep{Mccullagh}. How to perform selection in
log-linear models, or equivalently in binary graphical models,
depends upon the number of variables $p$. Indeed, the total number
of cell entries for a $p$-way contingency table is $2^p$, and the
total number of free parameters in the associated saturated model
is $2^p-1$. When $p$ is low, a standard approach for model
selection is greedy stepwise forward-selection or
backward-deletion: in each step, selection or deletion is based on
hypothesis testing at some level $\alpha$. However, the
computational complexity even for modestly dimensioned contingency
tables plus the multiple hypothesis testing issues related to such
a procedure has made it unpopular in this context. Consequently,
\cite{Dahinden} recently performed selection in log-linear models
by using an $\ell_1$-penalized version of the  log-likelihood,
extending the LASSO ideas \citep{Tibshirani1996} originally
designed for selection in linear models. However, computing the
(penalized) log-likelihood for log-linear models generally
requires the enumeration of each of the $2^p$ profiles, which is
not plausible for large $p$ (e.g., larger than about 30). For such
moderate-to-large values of $p$, alternative methods are required.
\vskip5pt

\noindent Roughly speaking, two approaches have been proposed in
the literature. First, exact inference can be performed in the
case of highly sparse models. For instance, exact computation via
the junction tree algorithm is manageable for highly sparse graphs
but becomes unwieldy for dense graphs \citep{SILee}. The second
approach is to use approximate inference. Notably, much attention
has recently been paid to methods relying on proxies for the exact
likelihood. \cite{Hofling} and \cite{Logitnet} proposed two
distinct algorithms to maximize an $\ell_1$-penalized version of
the so-called \emph{pseudo-likelihood} \citep{Besag}. These
methods are closely related to the one formerly proposed by
\cite{WainwrightLaffery} who used $\ell_1$-penalized logistic
regressions on each single node to construct the whole graphical
model. Besides these three methods, \cite{Banerjee} used a convex
relaxation technique to derive a Gaussian approximate
log-likelihood as well as its sparse maximum solution.\vskip5pt

\noindent Interestingly, \cite{Hofling} showed through an
extensive simulation study that approximate solutions (either
solutions maximizing the pseudo-likelihood or those derived from
the method proposed by \cite{WainwrightLaffery}) are much faster
and only slightly less accurate than exact methods. However, no
empirical evaluation of the Gaussian approximate solution proposed
by \cite{Banerjee} has ever been conducted and filling this gap is
the primary objective of this paper. We thereby propose to conduct
such an evaluation by comparing this method with the two other
approximate methods of \cite{WainwrightLaffery} and \cite{Hofling}
on simulated data. \vskip5pt

\noindent Here is a brief outline of the paper. In Section
\ref{Section_MM} we first summarize the principles of the
aforementioned approximate methods. We then present results from
an extensive empirical comparison study. A slight modification of
the method proposed by \cite{Banerjee} is especially shown to
achieve very good accuracy and to be extremely fast. Finally, we
present an application from a real example where we looked for
associations among causes of death in the database of French death
certificates of the year 2005 (Section \ref{Sec_Results}).

\section{Approximate methods for binary graphical models}\label{Section_MM}

\subsection{The Ising model}\label{Sec_binary_graph_models}

\noindent Let $\bX=(X^{(1)},...,X^{(p)}) \in \{0,1\}^p$ be a
$p$-dimensional vector of binary random variables. Given a random
sample $\bX_1,..., \bX_n$ of i.i.d. replicae of $\bX$, we wish to
study the associations between the coordinates of $\bX$. One way
to do so is to construct the  binary graphical model for the
random vector $\bX$, that is an undirected graph $\mathcal{G} =
(V,E)$, where $V$ contains $p$ vertices corresponding to the $p$
coordinates and the edges $E=(e_{k,\ell})_{1\leq k<\ell\leq p}$
describe the conditional independence relationships among
$X^{(1)},...,X^{(p)}$. The edge $e_{k,\ell}$ between $X^{(k)}$ and
$X^{(\ell)}$ is absent if and only if $X^{(k)}$ and $X^{(\ell)}$
are independent conditional on the other variables. For binary
graphical models, it is common to focus only on the family of
probability distributions given by the quadratic exponential
binary model \citep{WainwrightLaffery, Banerjee,
Hofling,Logitnet}, also known as the Ising model. Namely, for all
$\bx=(x^{(1)},...,x^{(p)}) \in \{0,1\}^p$, we assume that the
probability of $\bx$ is given by
\begin{equation}\label{Ising}
P(\bx,\Theta) = \exp\Big\{\sum_{k=1}^p\theta_kx^{(k)}+
\sum_{k=1}^{p-1}\sum_{\ell=k+1}^p\theta_{k,\ell}x^{(k)}x^{(\ell)}
- A(\Theta)\Big\},
\end{equation}
where the so-called $\log$ \emph{partition function} $A(\cdot)$ is
defined as follows
\begin{equation}\label{logpart}
A(\Theta) = \log
\sum_{\bx\in\{0,1\}^p}\exp\Big\{\sum_{k=1}^p\theta_kx^{(k)}
+\sum_{k=1}^{p-1}\sum_{\ell=k+1}^p\theta_{k,\ell}x^{(k)}x^{(\ell)}\Big\}.
\end{equation}
Note that $A(\cdot)$ is strictly convex and ensures that
$\sum_{\bx\in\{0,1\}^p} P(\bx,\Theta)=1$ (note also that the
strict convexity of this function ensures the identifiability of
the parameter matrix $\Theta$). From (\ref{Ising}), we have
\begin{equation}\label{OR_01}
\frac{P(X^{(k)}=1,X^{(\ell)}=1| X^{(j)}, j  \ne
k,\ell)/P(X^{(k)}=0,X^{(\ell)}=1| X^{(j)}, j \ne
k,\ell)}{P(X^{(k)}=1,X^{(\ell)}=0| X^{(j)}, j \ne
k,\ell)/P(X^{(k)}=0,X^{(\ell)}=0| X^{(j)}, j \ne
k,\ell)}=\exp(\theta_{k,\ell}).
\end{equation}
Therefore, the parameters $\theta_{k,\ell}$ are the conditional
log-odds ratios. The conditional independence between $X^{(k)}$
and $X^{(\ell)}$ is then equivalent to $\theta_{k,\ell}=0$, that
is, the edge $e_{k,\ell}$ is absent if and only if
$\theta_{k,\ell}=0$. Consequently, selection in binary graphical
models is equivalent to identifying the $(k,\ell)$ pairs for which
$\theta_{k,\ell}=0$. \vskip5pt

\noindent From (\ref{Ising}), using the fact that $x^{(k)}$ is
binary, and denoting by $\bbX=(\bX_1,...,\bX_n)^T$ the matrix
representing the whole dataset, the $\ell_1$-penalized
log-likelihood writes
\begin{equation}\label{Ising_likelihood}
l(\bbX,\Theta) = \sum_{\ell\geq k\geq 1}^p
(\bbX^T\bbX)_{k,\ell}\theta_{k,\ell} -
nA(\Theta)-n\lambda\|\Theta\|_1 ,
\end{equation}
where $\Theta$ is a symmetric matrix with $\theta_{kk}=\theta_k$
for $k=1,...,p$. However, because of the complexity of the
log-partition function, methods based on approximate inference are
needed in most cases and we recall the principle of three of them
in the following paragraphs.\vskip5pt

\noindent Let us begin by recalling that the approximation
established by \cite{Banerjee} is only valid for the
first-order-interaction log-linear model described above. On the
other hand, \cite{WainwrightLaffery}, \cite{Hofling} and
\cite{Logitnet} also only consider this simple model but
higher-order interaction models can be (at least theoretically)
handled with these methods, at a cost of a dramatically increased
computational time.

\subsubsection{Multiple logistic regressions}
\label{sec:ravikumar} From (\ref{Ising}), it is easy to see that,
for all $k=1,...,p$, setting
$\bx^{(-k)}=(x^{(1)},...,x^{(k-1)},x^{(k+1)},...,$ $x^{(p)})$, we
have
\begin{equation}
\mbox{logit} \{\P(X^{(k)}=1|\bx^{(-k)})\}=\ \sum_{\ell\neq
k}\theta_{k,\ell}x^{(\ell)}+\theta_k.\nonumber
\end{equation}
\cite{WainwrightLaffery} then extensively study a method (which
will be referred hereafter as \verb"SepLogit" following the
terminology adopted in \cite{Logitnet}) in which
$\ell_1$-penalized logistic regression is used to estimate the
neighborhood of each of the $p$ nodes in the graph separately.
\cite{WainwrightLaffery} gives rigorous consistency results in a
high-dimensional setting, where the number of nodes is allowed to
grow as a function of the number of samples. The authors give
sufficient conditions under which the method will consistently
estimate the neighborhood for every node in the graph
simultaneously. In a sense, the paper can be seen as a discrete
version of \cite{MeinBuhl}.\vskip5pt

\noindent For a finite number of samples, the $p$ logistic
regression problems are solved separately and, since the results
may be asymmetric, they can be combined in one of two ways to draw
a graph.  One possibility is to draw an edge between two nodes in
the graph only if each node is estimated to belong to the
neighborhood of the other (method \verb"SepLogit AND").
Alternatively, we can decide to draw an edge between two nodes so
long as at least one of them is estimated to belong to the
neighborhood of the other (method \verb"SepLogit OR").\vskip5pt

\noindent In our empirical comparison study, this method will be
implemented using the coordinate descent procedure developed by
\cite{glmnet} (and implemented in the \verb"glmnet" R package).

\subsubsection{Pseudo-likelihood maximization} \label{sec:wang}

One of the shortcomings of the method proposed by
\cite{WainwrightLaffery} is the aforementioned asymmetry. To
overcome this limitation, \cite{Hofling} and \cite{Logitnet}
recently proposed to use the pseudo-(log-)likelihood, first
suggested by \cite{Besag}. The pseudo-likelihood is formally
defined as
\begin{equation}
\sum_{i=1}^n\sum_{k=1}^p
\log\{\P(X_i^{(k)}|X_i^{(1)},...X_i^{(k-1)},X_i^{(k+1)},...,X_i^{(p)})\}.
\end{equation}
Accordingly, the approach based on the maximization of the
$\ell_1$-penalized pseudo-likelihood solves all $p$ logistic
regression problems simultaneously, while enforcing symmetry.
Apart from symmetry enforcement, this methods differs from the one
studied in \cite{WainwrightLaffery} in that the $\ell_1$-norm
penalty is applied to the entire network, while in \verb"SepLogit"
it is applied to each neighborhood. \vskip5pt

\noindent For future use, and still denoting by
$\bbX=(\bX_1,...,\bX_n)^T$ the matrix representing the whole
dataset, observe that the pseudo-likelihood writes
\begin{equation}\label{pseudo_likelihood_uncond}
\mbox{pseudo-}l(\bbX,\Theta)=-\sum_{i=1}^n\sum_{k=1}^p
\log\big\{1+\exp(-\tilde x_{i}^{(k)}\bbX^k[i,]\Theta[,k])\big\},
\end{equation}
where $\bbX^k$ is the same as $\bbX$ with $k$th column set to 1
and $\tilde x^{(k)}_{i} = 2x^{(k)}_{i}-1$ (i.e., $\tilde
x^{(k)}_{i}$ is the spin version of $x^{(k)}_{i}$). Here and
elsewhere, $M[i,]$ (resp. $M[,k]$) denotes the $i$-th row (resp.
$k$-th column) of a matrix $M$. \vskip5pt

\noindent \cite{Hofling} first develop and implement an algorithm
for maximizing the $\ell_1$-penalized pseudo-likelihood function,
using a local quadratic approximation to the pseudo-likelihood.
They then use this algorithm as a building block for a new
algorithm that maximizes the true log-likelihood. However, as we
already said, they observed that the approximate pseudo-likelihood
is much faster than the exact procedure, and only slightly less
accurate. Therefore, to save computational time, we only
considered the approximate pseudo-likelihood in this paper. In the
forthcoming empirical comparison study, this method will be
implemented using the \verb"BMN" R package and will be referred to
as \verb"BMNPseudo". \vskip5pt

\noindent Interestingly, and as pointed out by \cite{Hofling}, the
derivative of the pseudo-likelihood on the off-diagonal is roughly
twice as large as the derivative of the exact likelihood.
Moreover, in the case $p=2$, it is easy to see that the deviance
of the model with no association (i.e. minus twice the difference
between the log-likelihood of this model and the log-likelihood of
the saturated model) when computing with the pseudo-likelihood is
twice as large as the one computed with the exact likelihood
(while, obviously, the pseudo-likelihood coincides with the exact
likelihood for the model with no interaction). The generalization
of this striking result for higher $p$ is not straightforward, but
our empirical examples suggest it may hold at least approximately
(see Section \ref{Sec_Comp_Approx_Deviances}). Therefore, we will
consider methods relying on both $\mbox{pseudo-}l(\bbX,\Theta)$
(method \verb"BMNPseudo") and $\mbox{pseudo-}l(\bbX,\Theta)/2$
(method \verb"BMNPseudo 1/2") in the sequel. \vskip5pt\vskip5pt

\noindent For the sake of completeness, we shall add that
\cite{Logitnet} develop a gradient-descent algorithm to maximize
the $\ell_1$-penalized pseudo-likelihood. They further propose an
extension to account for spatial correlation among the variables
(which was relevant in their example dealing with genomic data).
However, the corresponding \verb"LogitNet" R package was not
available at the time we wrote this paper, so we were not able to
include it in our empirical comparison study.\vskip5pt

\subsubsection{Gaussian Approximation of the Ising log-likelihood}
The basic idea of the method described by \cite{Banerjee} is to
replace the log-partition function in the Ising model with an
upper bound suggested by \cite{WainwrightJordan}.  The resulting
approximation can then, with some manipulation, be put in a form
that can be solved efficiently using block coordinate descent. In
order to add some specific details, we shall define some notation.
Denote by $(\bZ_{1},...,\bZ_{n}) \in \{-1,1\}^{p\times n}$ the
spin version of $(\bX_{1},...,\bX_{n})$, and let
$\overline{Z}^{(k)}$ denote the sample mean of variable $Z^{(k)}$,
for $k=1,...,p$. \noindent Now, define the empirical covariance
matrix $S$ as
\begin{equation}
S = \frac{1}{n}\sum_{i=1}^n(\bZ_{i}-\overline Z)(\bZ_{i}-\overline
Z)^T,
\end{equation}
where $\overline Z$ is the vector of sample means
$\overline{Z}^{(k)}$.  Making use of a \emph{convex relaxation}
and a useful upper bound on the $\log$-partition function obtained
by \cite{WainwrightJordan}, \cite{Banerjee} established that an
approximate sparse maximum likelihood solution for a given
$\lambda$ has the following form
\begin{eqnarray}
\hat \theta^\lambda_k &=& \overline{Z}^{(k)}, \nonumber \\
\hat \theta^\lambda_{k,\ell} &=&-(\hat
\Sigma_\lambda^{-1})_{k\ell}, \label{sol_opt_binary}
\end{eqnarray}
where the matrix $\hat\Sigma_\lambda^{-1}$ is the solution of the
following optimization problem
\begin{equation}\label{optim_binary_0}
\hat \Sigma_\lambda^{-1}= \mbox{arg}\max_M \big\{ \log|M|-
\mbox{tr}(M(S+\mbox{diag}(1/3)))-\lambda\|M\|_1\big\}.
\end{equation}
More precisely, \cite{Banerjee} proposed a block-coordinate
descent algorithm to solve a dual formulation of
(\ref{optim_binary_0}), which can be written as
\begin{equation}\label{optim_binary}
\hat \Sigma_\lambda = \mbox{arg}\max_W\big\{\log|W| :
W_{kk}=S_{kk}+\frac{1}{3},|W_{k\ell}-S_{k\ell}|\leq\lambda \big\}.
\end{equation}
\noindent In the Gaussian case, \cite{Banerjee} showed  that the
$\ell_1$-penalized covariance selection problem could be written
\begin{equation}\label{dual_1}
\widehat \Sigma = \mbox{arg}\max_W\big\{\log|W| :
\|W-S_G\|_\infty\leq\lambda \big\},
\end{equation}
where $S_G$ is the empirical covariance matrix attached to a given
sample  of Gaussian vectors. An algorithm for handling binary
graphical models can be derived by comparing (\ref{optim_binary})
and (\ref{dual_1}). The original $\{0,1\}$ data has first to be
transformed into $\{-1,1\}$ data. Then, adding the constant $1/3$
to the diagonal elements of the resulting empirical covariance
matrix, the algorithms developed in the Gaussian case (in
particular the \verb"glasso" R package developed by
\cite{FriedmanHastieTib}) can be reused.\vskip5pt

\noindent A common question when working with Gaussian variables
is whether to standardize them, or equivalently, whether to use
the covariance or the correlation matrix. Moreover, in the binary
case, the correlation coefficient between two variables (also
known as the $\phi$-coefficient) is closely related to the
$\chi^2$ statistic used to test for (marginal) independence.
Putting these two observations together, we decided to evaluate a
simple modification of the method proposed by \cite{Banerjee}
where the quantity $S$+diag(1/3) is replaced by the correlation
matrix. Lastly, we also decided to evaluate the modification in
which $S$+diag(1/3) is simply replaced by $S$. \vskip5pt

\noindent To recap, we will consider the three following
optimization problems
\begin{equation}
\widehat C^\nu_{\lambda}= \mbox{arg}\max_{M} \big\{\log|M| -
\mbox{tr}(MS^\nu) -\lambda\|M\|_1 \big\} \mbox{, for
}\nu=1,2,3,\label{Gauss_Approx_sol}
\end{equation}
where $S^1=(\mbox{Cov}(\bZ)+\mbox{diag}(1/3))$, $
S^2=\mbox{Cov}(\bZ)$ and $ S^3=\mbox{Cor}(\bZ)$. For any
$\lambda$, and every $\nu=1,2,3$ an estimation of
$\theta_{k,\ell}$ is then given by $-(\widehat
C^\nu_{\lambda})_{k\ell}$.

\noindent In our empirical comparison study, the three methods
will be implemented using the \verb"glasso" R package
\citep{FriedmanHastieTib} and will be referred to as
\verb"GaussCov 1/3", \verb"GaussCov" and \verb"GaussCor" for the
choices $\nu=1$, $\nu=2$ and $\nu=3$ respectively (we may as well
use the generic expression \verb"GaussApprox" when dealing with
either methods).

\subsection{Sparsity
parameter selection}\label{Sec_sel_sparsity_param} Two procedures
for selecting tuning parameters are generally considered, namely
cross-validation (CV) and Bayesian Information Criterion (BIC),
the latter being computationally more efficient as suggested by
\cite{YuanLin} for instance. In the case of Gaussian graphical
models, \cite{Gaoetal} further demonstrate the advantageous
performance of BIC for sparsity parameter selection through
simulation studies. In this paper, we therefore decided to only
consider BIC.\vskip5pt

\noindent When trying to select the optimal sparsity parameter
$\lambda$ using either CV or BIC, however, one has to pay
attention to the following fact. Since, for each $\lambda>0$,
estimates of the parameters of interest are shrunk, using them for
choosing $\lambda$ from CV or BIC often results in severe
over-fitting \citep{LARS}. Therefore, un-shrunk estimates have to
be derived before computing the BIC. \vskip5pt

\noindent Taking the example of methods \verb"GaussApprox", for
any $\lambda$, we have to compute the un-shrunk matrix
\begin{equation}\label{pr_unshrunk_gauss}
\widetilde C^\nu_{\lambda} =
\mbox{arg}\!\!\!\max_{M\in\mathcal{M}_{\lambda}^+} \big\{\log|M| -
\mbox{tr}(MS^\nu)\big\}.
\end{equation}
Here, $\mathcal{M}_{\lambda}^+=\{M\succ0: M_{k,\ell}=0 \mbox{ for
couples $(k,\ell)$ such that } (\widehat
C^\nu_{\lambda})_{k,\ell}=0\}$ ($\{M\succ0\}$ being the set of
positive definite matrices of order $p$, and $\widehat
C^\nu_{\lambda}$ being as in (\ref{sol_opt_binary})). To solve the
optimization problem (\ref{pr_unshrunk_gauss}), one approach is to
reuse the algorithm used to solve (\ref{Gauss_Approx_sol}) after
replacing the scalar parameter $\lambda$ by the penalty matrix
$\Lambda$ such that $\Lambda_{k,\ell}=0$ if
$\hat\theta^\lambda_{k,\ell}\neq0$, and $\Lambda_{k,\ell}=\infty$
otherwise, where $\hat\theta^\lambda_{k,\ell}$ is the shrunk
estimation of the coefficient $\theta_{k,\ell}$ obtained with the
value $\lambda$ and is used as an initial value for the
optimization. Alternative approaches might be considered, such as
the algorithm developed by \cite{DahlVandenberghe} for instance.
\vskip5pt

\noindent Defining
$$\mathcal{L}^\nu_\lambda= \log|\widetilde C^\nu_{\lambda}|-
\mbox{tr}(\widetilde C^\nu_{\lambda}S^\nu) ,$$ the BIC procedure
now simply corresponds to selecting the sparsity parameter
$\lambda^\nu_{{\footnotesize\mbox{BIC}}}$ such that
\begin{equation}\label{BIC}
\lambda^\nu_{{\footnotesize\mbox{BIC}}} = \arg\max_\lambda
\big\{n\mathcal{L}^\nu_\lambda - K^\nu_\lambda \log(n) \big\},
\end{equation}
where $K^\nu_\lambda=\sum_{k\geq \ell} \1\{(\widehat
C^\nu_{\lambda})_{k,\ell}\neq 0\}$ is the degree of freedom of the
model selected with the sparsity parameter $\lambda$
\citep{YuanLin}. \vskip5pt

\subsection{Estimation of the conditional odds-ratios}\label{Sec_estim_odds_ratio}

In the binary case, a standard measure of the strength of
association between two variables is the (conditional) odds-ratio,
which is related to coefficient $\theta_{k,\ell}$ (see
(\ref{OR_01})). Therefore, consistent estimates of the parameters
$\theta_{k,\ell}$ would yield consistent estimates of the
conditional odds-ratios. Here again un-shrunk estimates are
preferable, and the methods described in the previous section have
to be used.

\section{Simulation study}\label{Sec_Simul}

In this section, we compare the model selection performances as
well as the computational time for the methods described in the
previous Section. Results are presented for $p=10$ and $p=50$. The
choice $p=10$ has been made for several reasons. First, for such
low values of $p$ the true log-likelihood can be quickly computed,
and we can then compare it with the approximate log-likelihoods
(see Section \ref{Sec_Comp_Approx_Deviances} below). Second,
approximate methods are still faster to compute when $p$ is small,
and conclusions drawn in the case of low $p$ are likely to hold
for high $p$ as well (as will be confirmed from our
results).\vskip5pt

\subsection{Evaluation criteria}\label{Sec_Evaluation_Criteria}

For each method, every value of $\lambda$ corresponds to a
sparsity structure for the matrix $\Theta$ that can be compared
with the true sparsity structure. Namely, for all $\lambda$ and
for each method, we can compute the rate of true positives
(correctly identified associations), the rate of false positives
(incorrectly identified associations) as well as the overall
accuracy. Precision and recall (the latter being identical to the
true positive rate) can also be computed, as well as their
harmonic mean, often referred to as the F1-score. \vskip5pt

\noindent In a first evaluation study, we present for each method
the performances achieved by the "oracle" model, that is the model
constructed with optimal sparsity parameter regarding accuracy.
Such an evaluation was not conducted for \verb"SepLogit" because
under this method the $\ell_1$ penalty is applied to each
neighborhood and the "oracle" model would invariably coincide with
the true model. The alternative would be to force the algorithm to
choose the same sparsity parameter value for every regression
model. However, using this alternative approach, we sometimes
obtained "oracle" models that achieved performances worse than the
models selected by the BIC procedure. Therefore, we do not
recommend to force the algorithm to choose the same sparsity
parameter value for every regression model. \vskip5pt

\noindent In a second evaluation study, we present for each method
the performances achieved by the model selected according the BIC
procedure described above. For methods \verb"GaussApprox",
un-shrunk estimates were derived along the lines described in
Section \ref{Sec_sel_sparsity_param}. A similar approach was used
for method \verb"BMNPseudo".  The \verb"BMN" R package also allows
the use of a  matrix of penalty coefficients. For \verb"SepLogit",
we had to slightly adapt this approach because the \verb"glmnet"
package does not allow for building models with only un-penalized
coefficients. So, whenever needed, a standard logistic regression
model was used to get un-shrunk estimates. This may make the
method a little slower, but not much since for each variable, this
situation can only arise for the smallest tested $\lambda$ value,
and only if this smallest tested $\lambda$ value corresponds to
the saturated model. \vskip5pt

\noindent In addition, the computational time is reported. More
precisely, we used a grid
$[\lambda^{\footnotesize{\mbox{min}}}:=\lambda^{\footnotesize{\mbox{max}}}/1000,
..., \lambda^{\footnotesize{\mbox{max}}}]$ of 50 equally-spaced
values (on a log-scale) for the parameter $\lambda$ and we report
the time needed to compute the 50 corresponding models for each
method ($\lambda^{\footnotesize{\mbox{max}}}$ is the data derived
smallest value for which all coefficients are zero). Each method
was run on a Windows Vista machine with Intel Core 2DUO 2.26GHz
with 4GB RAM in the case $p=10$ and on a MAC Pro machine with
intel Xeon 2$\times$2.26GHz Quad Core with 6GB RAM in the case
$p=50$ (the MAC Pro machine was approximately 3.5 times as fast as
the Windows machine).

\subsection{Data generation}

\subsubsection{The case $p=10$}

In model (\ref{Ising}), given that $n$ individuals are observed,
the distribution of the corresponding cell counts ${\bf
n}=(n_{\bx}, \bx\in\{-1,1\}^p)$ is multinomial with probability
${\bf P}=(P(\bx,\Theta), \bx\in\{-1,1\}^p)$. Accordingly, given a
symmetric matrix $\Theta$, data were drawn from the multinomial
distribution with probability vector ${\bf P}$. Four matrices
$\Theta^{(1)}$, $\Theta^{(2)}$, $\Theta^{(3)}$ and $\Theta^{(4)}$
were considered, leading to four different simulation designs.

\noindent For $\Theta^{(1)}$, "primary" coefficients
$\theta_{k,\ell}$ were simulated independently using a normal
distribution with mean zero and variance $\sigma$ for some
$\sigma>0$. Subsequently, only coefficients $\theta_{k,\ell}$ with
an absolute value greater than 0.06 (corresponding to a
conditional odds-ratio of $\exp(4\times0.06)\simeq 1.27$, since
for $\{-1,1\}$ variables, the conditional odds-ratio is
$\exp(4\theta_{k,\ell})$) were retained, all others being set to
0. The function $A(\Theta^{(1)})$ was then computed according to
Equation (\ref{logpart}). Selecting $\sigma=0.05$ led to a true
model with 10 associations (among the $p(p-1)/2=45$ potential
associations). \vskip5pt

\noindent Matrices $\Theta^{(2)}$ and $\Theta^{(3)}$ were
constructed so that they share the same sparsity pattern as
$\Theta^{(1)}$, i.e.,
$$\{(k,\ell):\theta^{(1)}_{k,\ell}=0\}=\{(k,\ell):\theta^{(2)}_{k,\ell}=0\}=\{(k,\ell):\theta^{(3)}_{k,\ell}=0\}$$
but they have different non-zero coefficients. In either cases, we
selected $(\theta_{1,1},...,\theta_{10,10})=(-1.3,...,0)$. For
matrix $\Theta^{(2)}$, the non-zero $\theta_{k,\ell}$ coefficients
were set to $\pm 0.2$, while they were set to $\pm 0.4$ for matrix
$\Theta^{(3)}$.\vskip5pt

\noindent  For matrix $\Theta^{(4)}$, we proceeded as for matrix
$\Theta^{(1)}$ but we selected $\sigma=0.3$ and only the
$\theta_{k,\ell}$ coefficients with an absolute value greater than
0.2 were retained (the others being set to 0). Moreover, we
selected $(\theta_{1,1},...,\theta_{10,10})=(-1.8,...,0)$. This
led to a true model with 19 associations. \vskip5pt

\begin{figure}
\begin{center}
\includegraphics[width=0.35\textwidth]{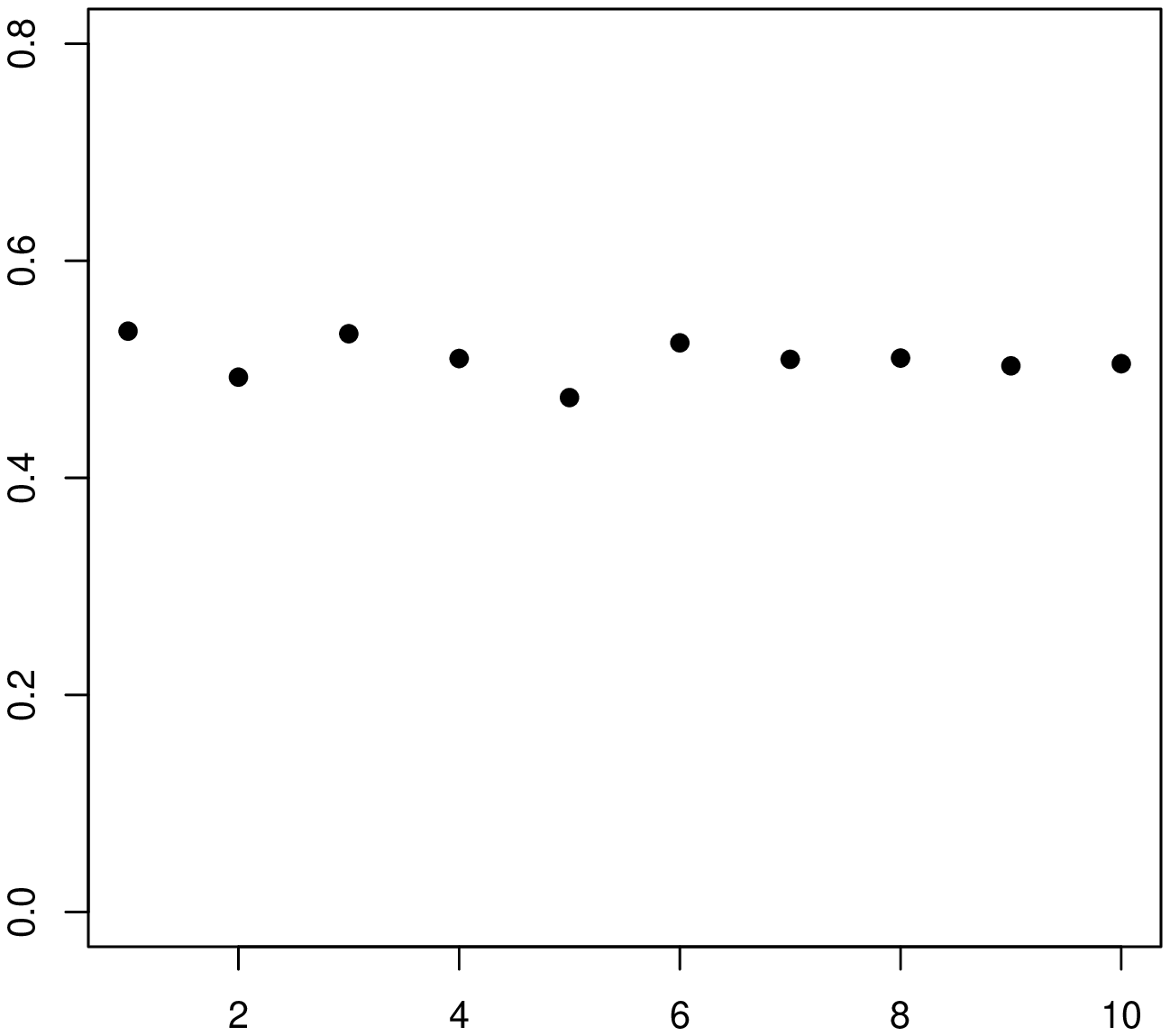}
\hspace{.35in}
\includegraphics[width=0.35\textwidth]{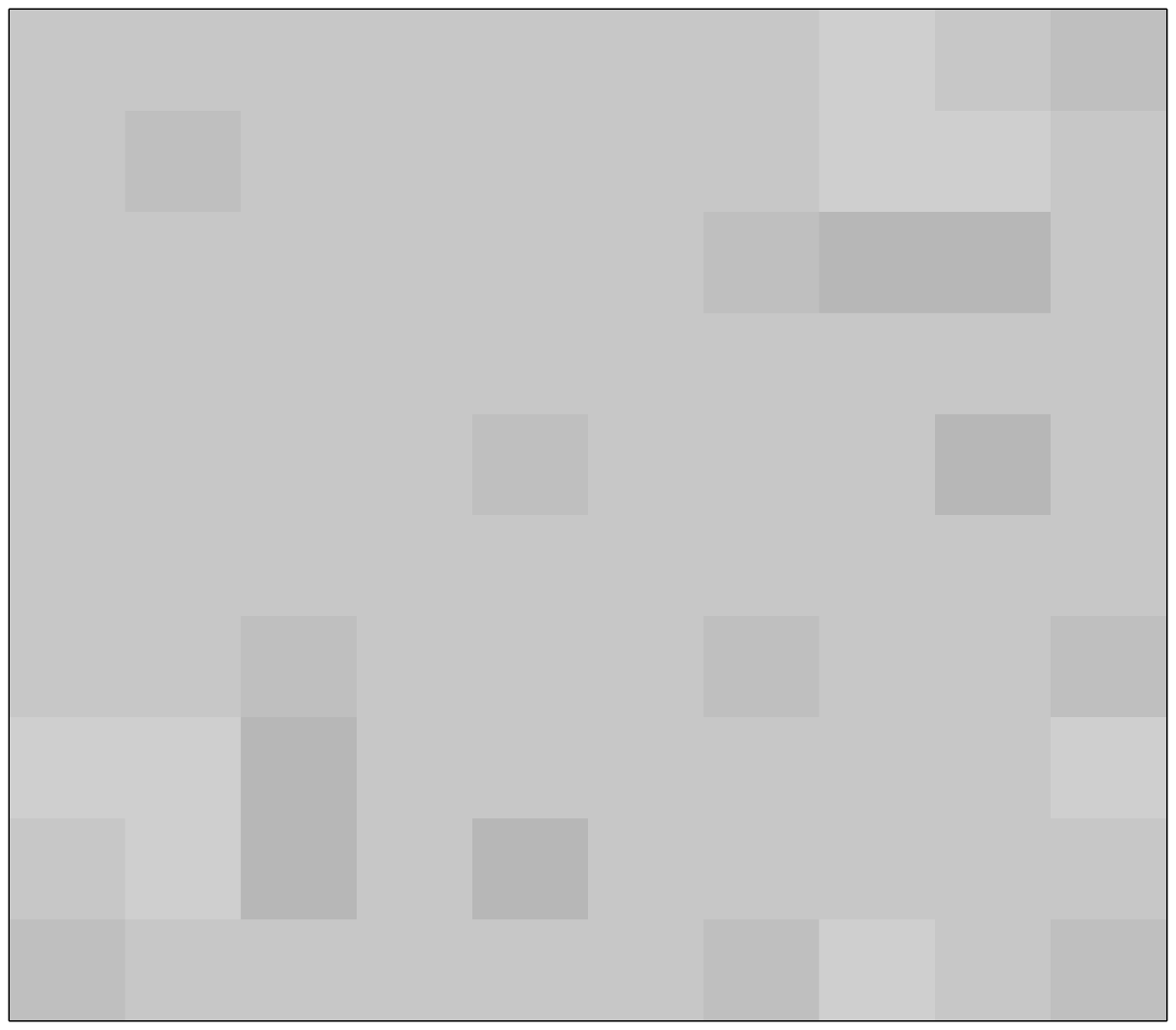}
\vspace{.05in}
\includegraphics[width=0.35\textwidth]{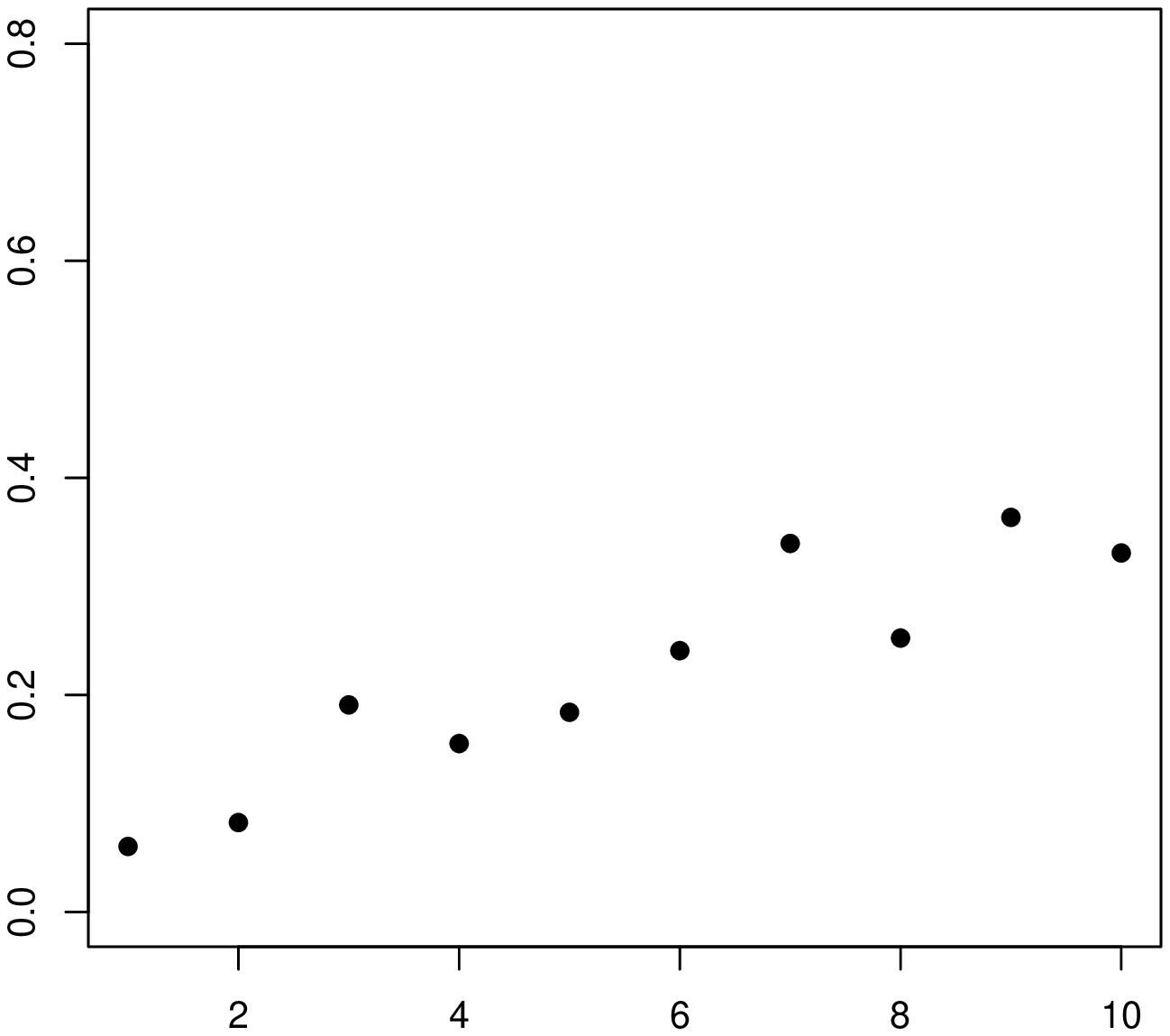}
\hspace{.35in}
\includegraphics[width=0.35\textwidth]{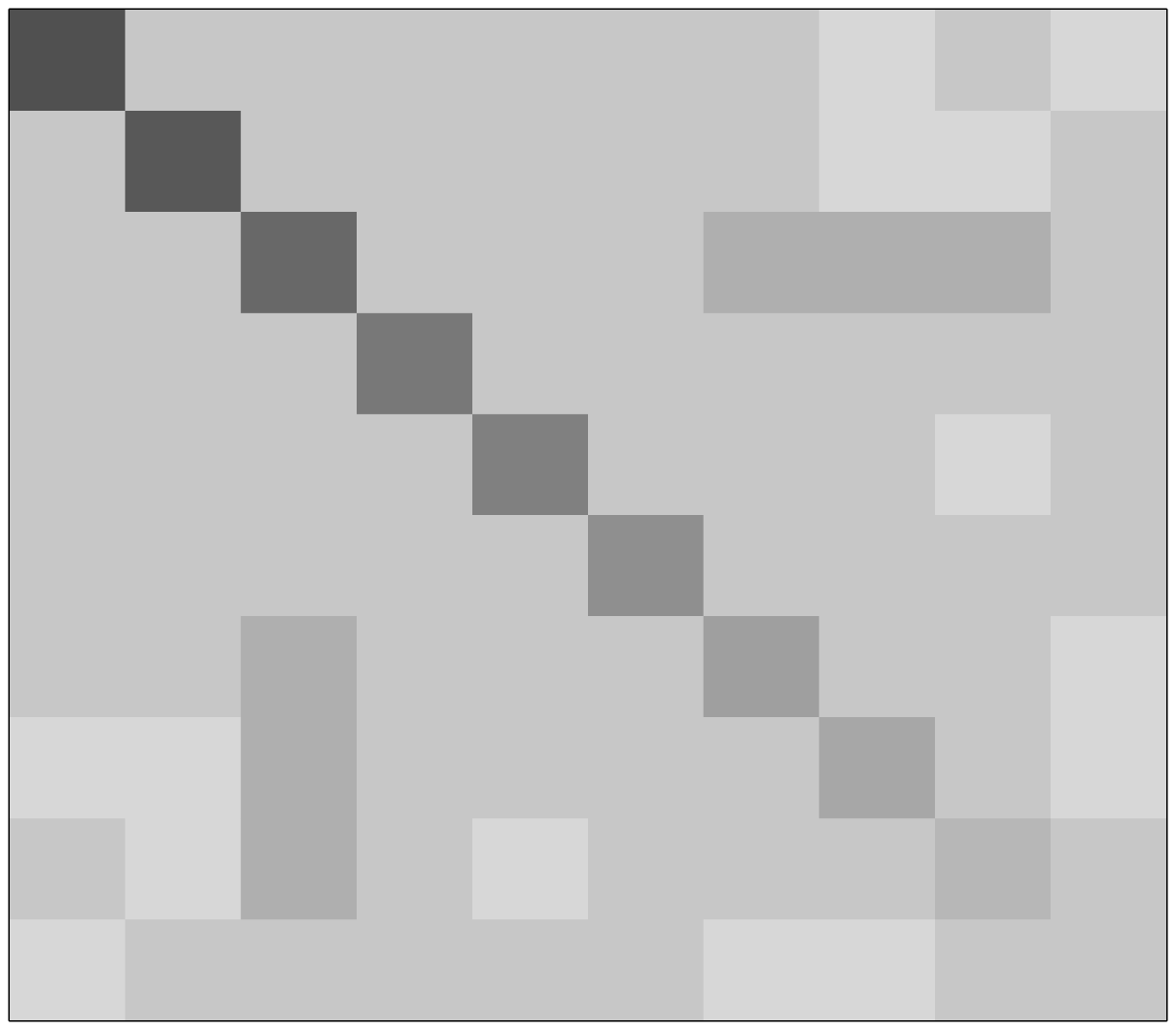}
\vspace{.05in}
\includegraphics[width=0.35\textwidth]{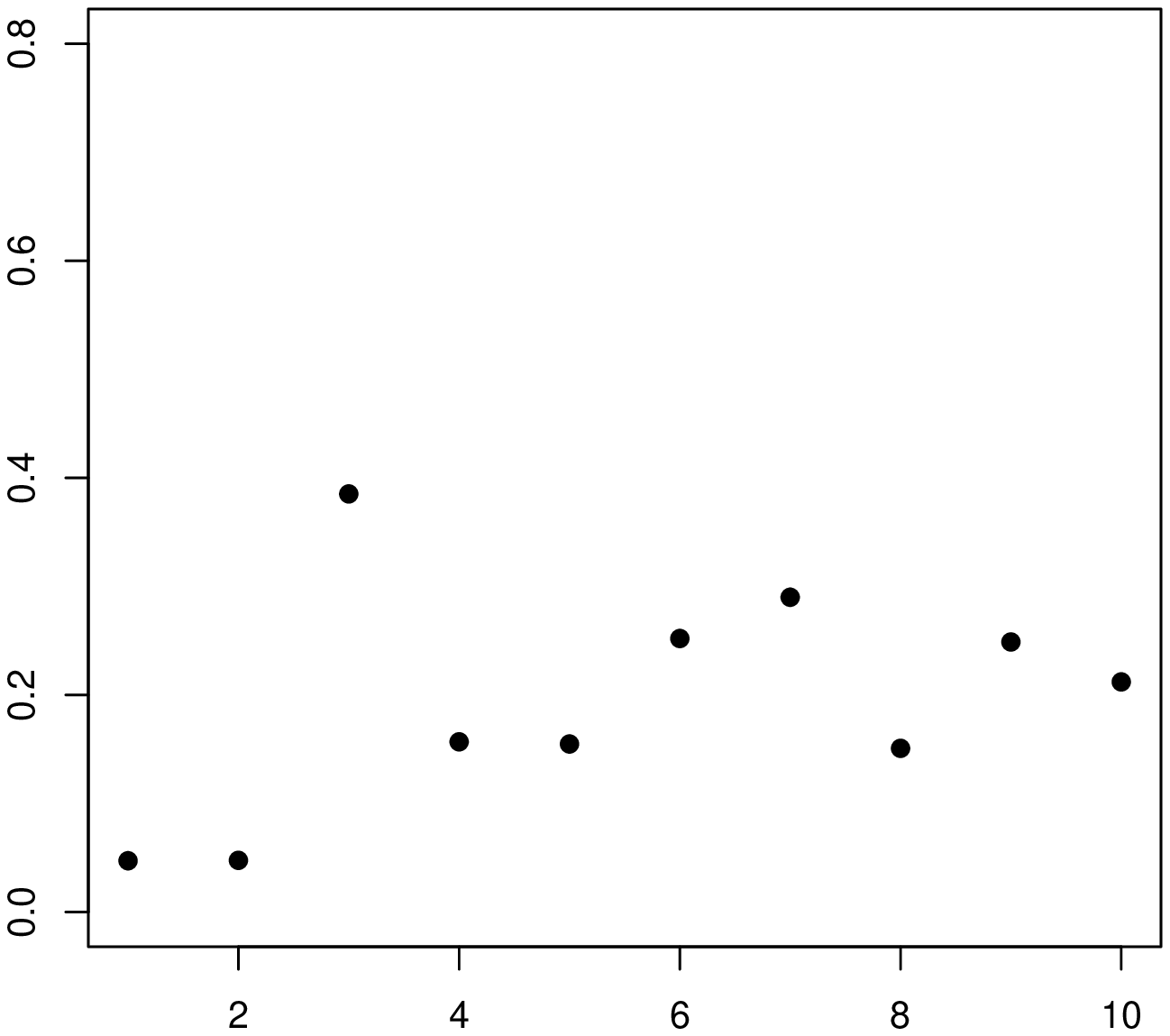}
\hspace{.35in}
\includegraphics[width=0.35\textwidth]{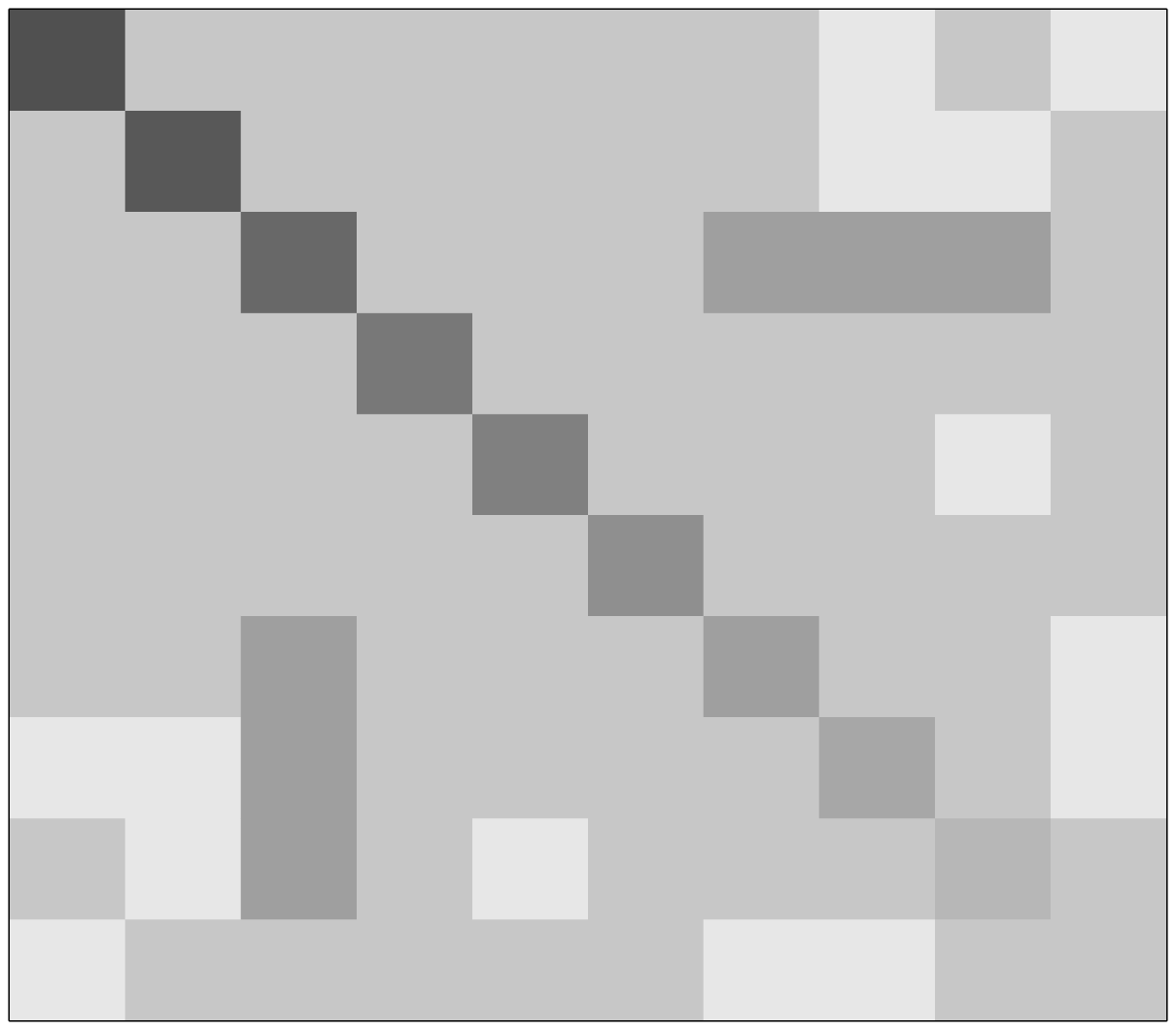}
\vspace{.05in}
\includegraphics[width=0.35\textwidth]{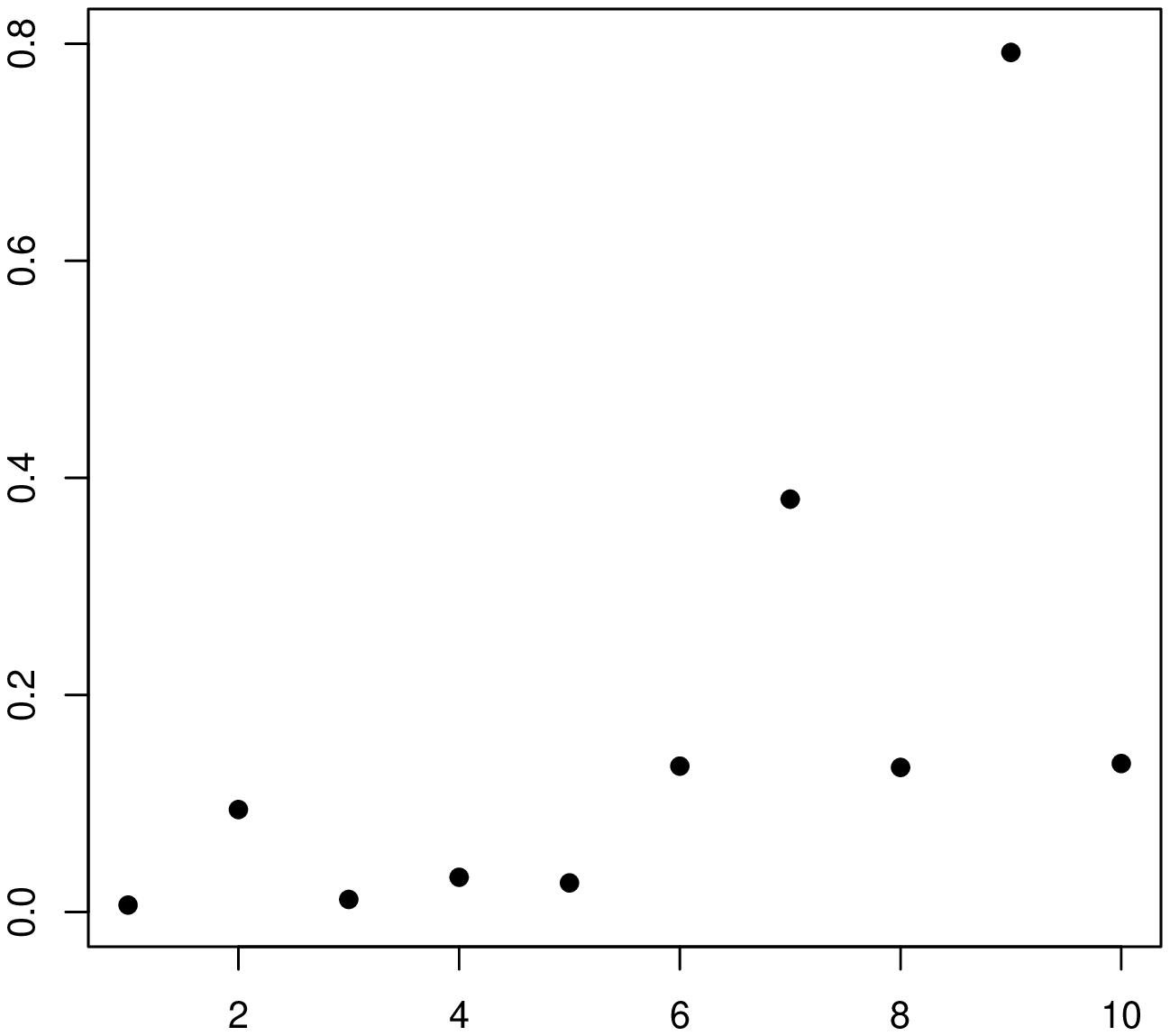}
\hspace{.35in}
\includegraphics[width=0.35\textwidth]{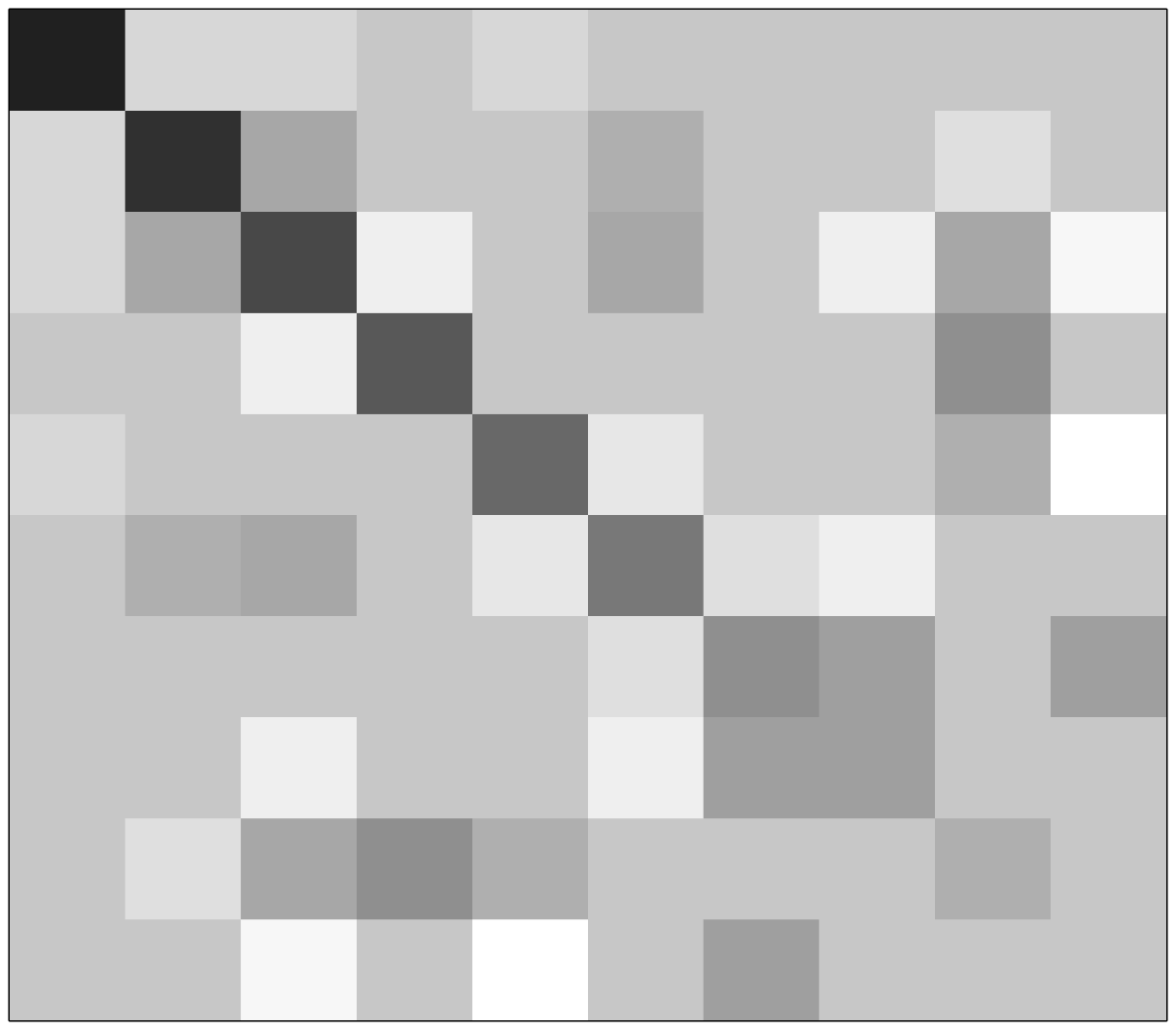}
\end{center}
\caption{Marginal probabilities $\P(X^{(k)}=1)$, for $k=1,...,10$,
estimated on a sample of size $n=2500$ generated using the matrix
$\Theta=\Theta^{(j)}, j=1,2,3,4$ are presented in the left panels.
A graphical representation of matrix $\Theta^{(j)}, j=1,2,3,4$ is
given in the right panels. }\label{Fig_Descr_Data_p10}
\end{figure}

\noindent A graphical representation of matrices $\Theta^{(1)}$,
$\Theta^{(2)}$, $\Theta^{(3)}$ and $\Theta^{(4)}$ as well as the
corresponding marginal probabilities $\P(X^{(k)}=1)$, for
$k=1,...,10$, estimated on a sample of size $n=2500$ are presented
on Figure \ref{Fig_Descr_Data_p10}.

\subsubsection{The case $p=50$} For $p=50$, we first considered the
case of block-diagonal matrices $\Theta$. For $j=1,2,3,4$, we then
used matrices $\Theta$ of the form
$\mbox{diag}(\Theta^{j},\Theta^{j},\Theta^{j},\Theta^{j},\Theta^{j})$.\vskip5pt

\noindent In a fifth example, matrix $\Theta^{(5)}$ was build as
follows. For every $k>\ell\geq1$, we first draw one observation
$u$ from a (0,1)-uniform distribution, and $\theta^{(5)}_{k,\ell}$
was then set to 0 if $u<0.9$, $\log(2)$ if $u\geq 0.95$ and
$\log(1.5)$ otherwise. The resulting true model consisted of 125
edges. Coefficients $\theta^{(5)}_{k,k}$ were set to
(logit(0.1),...,logit(0.2)). Gibbs sampling was further used to
generate the data (consisting of $\{0,1\}$ variables this time).

\begin{figure}
\begin{center}
\includegraphics[width=0.35\textwidth]{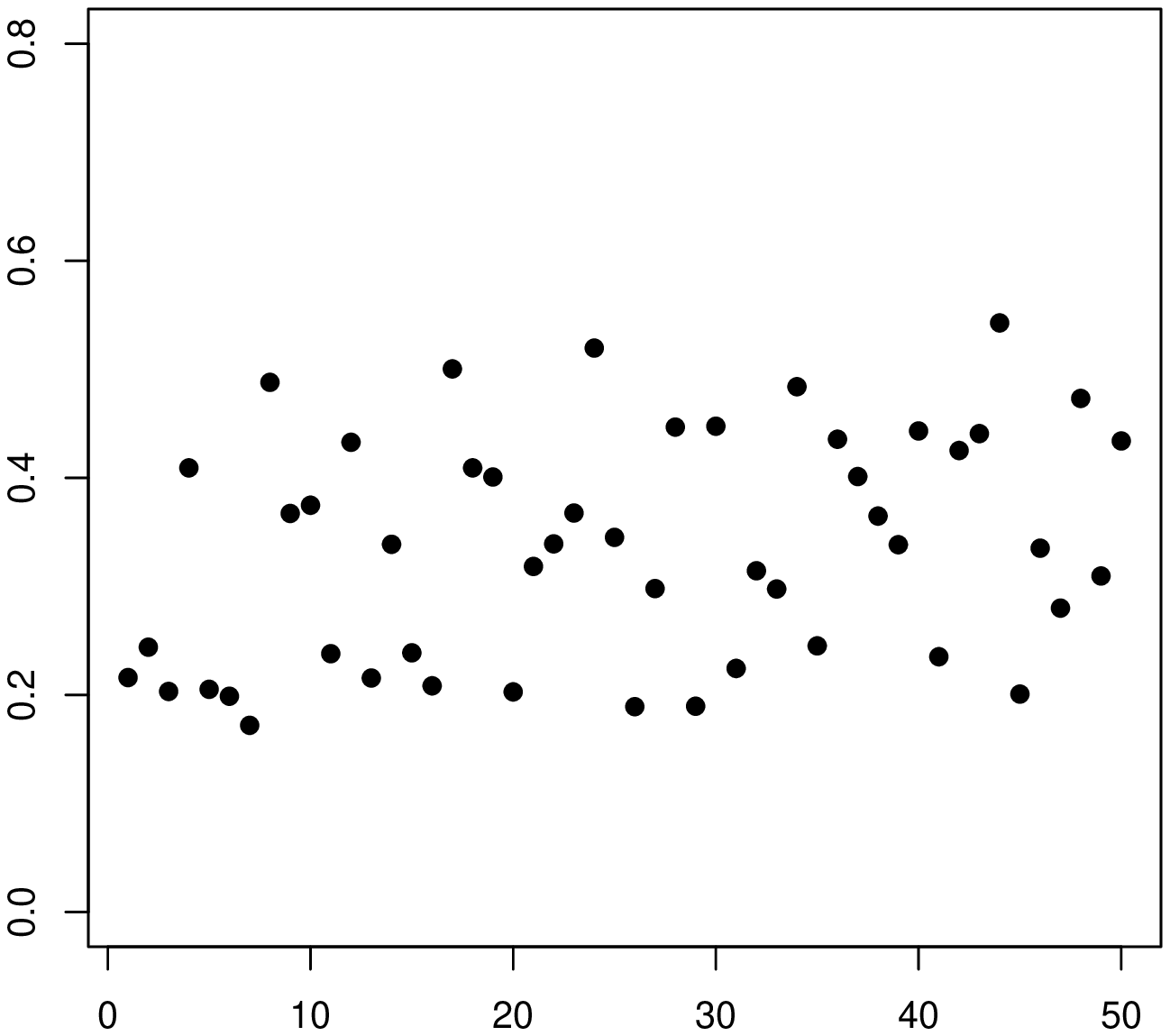}
\hspace{.35in}
\includegraphics[width=0.35\textwidth]{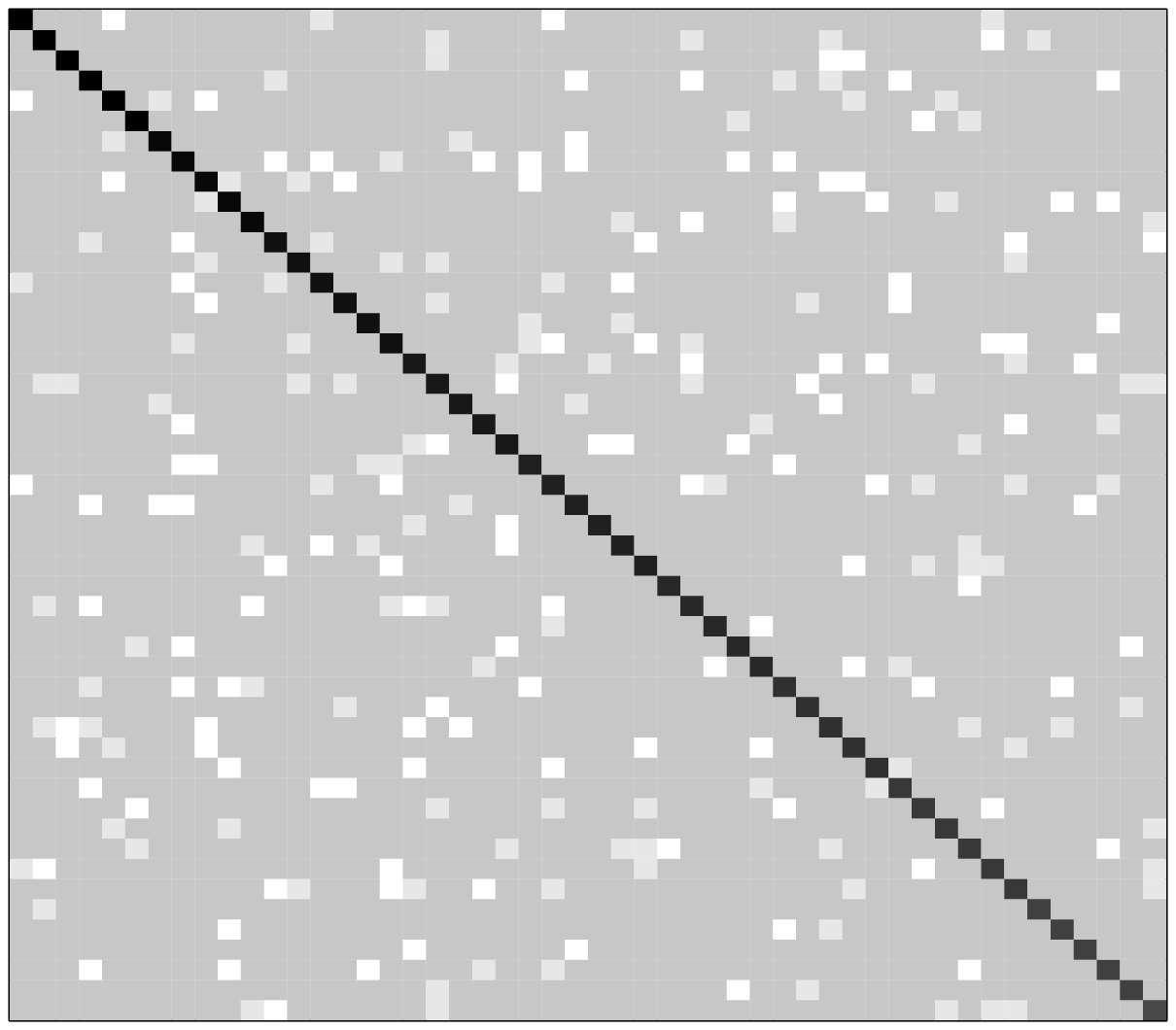}
\end{center}
\caption{Marginal probabilities $\P(X^{(k)}=1)$, for $k=1,...,50$,
estimated on a sample of size $n=2500$ generated using the matrix
$\Theta=\Theta^{(5)}$ are presented in the left panel. A graphical
representation of matrix $\Theta^{(5)}$ is given in the right
panel.}\label{Fig_Descr_Data}
\end{figure}

\subsection{Empirical comparison of the approximate
deviances}\label{Sec_Comp_Approx_Deviances} In this section, our
goal is to empirically evaluate the approximate likelihoods on
which the methods under study rely. To do so, we will focus on the
case where $p=10$ since the exact log-likelihood of the Poisson
log-linear model can be computed for such a value of $p$. For each
of the four $\Theta$ matrices described above, we proceed as
follows. We generate a random sample of size $n=500$, and for each
value of the tuning parameter $\lambda$ on an appropriate grid, we
apply method \verb"BMNPseudo". This leads to some sparsity
structure in the corresponding Ising model and we can then compute
the Gaussian approximate log-likelihoods
$\mathcal{L}^{{\footnotesize\mbox{G$_\nu$}}}_\lambda$, $\nu=1,2,3$
(see (\ref{pr_dev_gauss}) below) as well as both the exact Poisson
log-linear log-likelihood and the pseudo-likelihood for the Ising
model corresponding to this particular sparsity structure. More
precisely, the following quantities were considered,
\begin{eqnarray}
\mathcal{L}^{{\footnotesize\mbox{Ps}}}_\lambda&=&-\sum_{i=1}^n\sum_{k=1}^p
\log\big\{1+\exp(-\tilde
x_{i}^{(k)}\bbX^k[i,]\Theta_\lambda[,k])\big\} \label{pr_dev_pseudo}\\
\mathcal{L}^{{\footnotesize\mbox{Po}}}_\lambda&=&\sum_{i=1}^n
\log\big\{P(\bx_i,\Theta_\lambda^{\footnotesize\mbox{Po}})\big\}\label{pr_dev_poisson}\\
\mathcal{L}^{{\footnotesize\mbox{G$_\nu$}}}_\lambda&=& \log|
C^\nu_{\lambda}|- \mbox{tr}(C^\nu_{\lambda}S_\nu)\mbox{ for
}\nu=1,2,3.\label{pr_dev_gauss}
\end{eqnarray}
In (\ref{pr_dev_pseudo}), $\Theta_\lambda$ stands for the
un-shrunk matrix derived under the sparsity structure inferred
from method \verb"BMNPseudo" with the sparsity parameter value
$\lambda$, and $\tilde x_{i}^{(k)}$ and $\bbX^k$ are as in
(\ref{pseudo_likelihood_uncond}). In (\ref{pr_dev_poisson}),
$\Theta_\lambda^{\footnotesize\mbox{Po}}$ is the matrix of
coefficients obtained using a Poisson log-linear model under the
constrained induced by this sparsity structure, and
$P(\bx,\Theta)$ is as in (\ref{Ising}). Lastly, in
(\ref{pr_dev_gauss}), $S_1=(\mbox{Cov}(\bZ)+\mbox{diag}(1/3))$,
$S_2=\mbox{Cov}(\bZ)$, $S_3=\mbox{Cor}(\bZ)$ and $C^\nu_{\lambda}$
is defined as
$$C^\nu_{\lambda}= \mbox{arg}\max_{M\in\mathcal{M}_\lambda^+} \big\{\log|M| - \mbox{tr}(M
S_\nu)\big\} ,$$ where $\mathcal{M}_{\lambda}^+=\{M\succ0:
M_{k,\ell}=0 \mbox{ for couples $(k,\ell)$ such that }
(\Theta_{\lambda})_{k,\ell}=0\}$ (with $\Theta_\lambda$ as in
(\ref{pr_dev_pseudo})). \vskip5pt

\begin{figure}
\begin{center}
\includegraphics[width=0.47\textwidth]{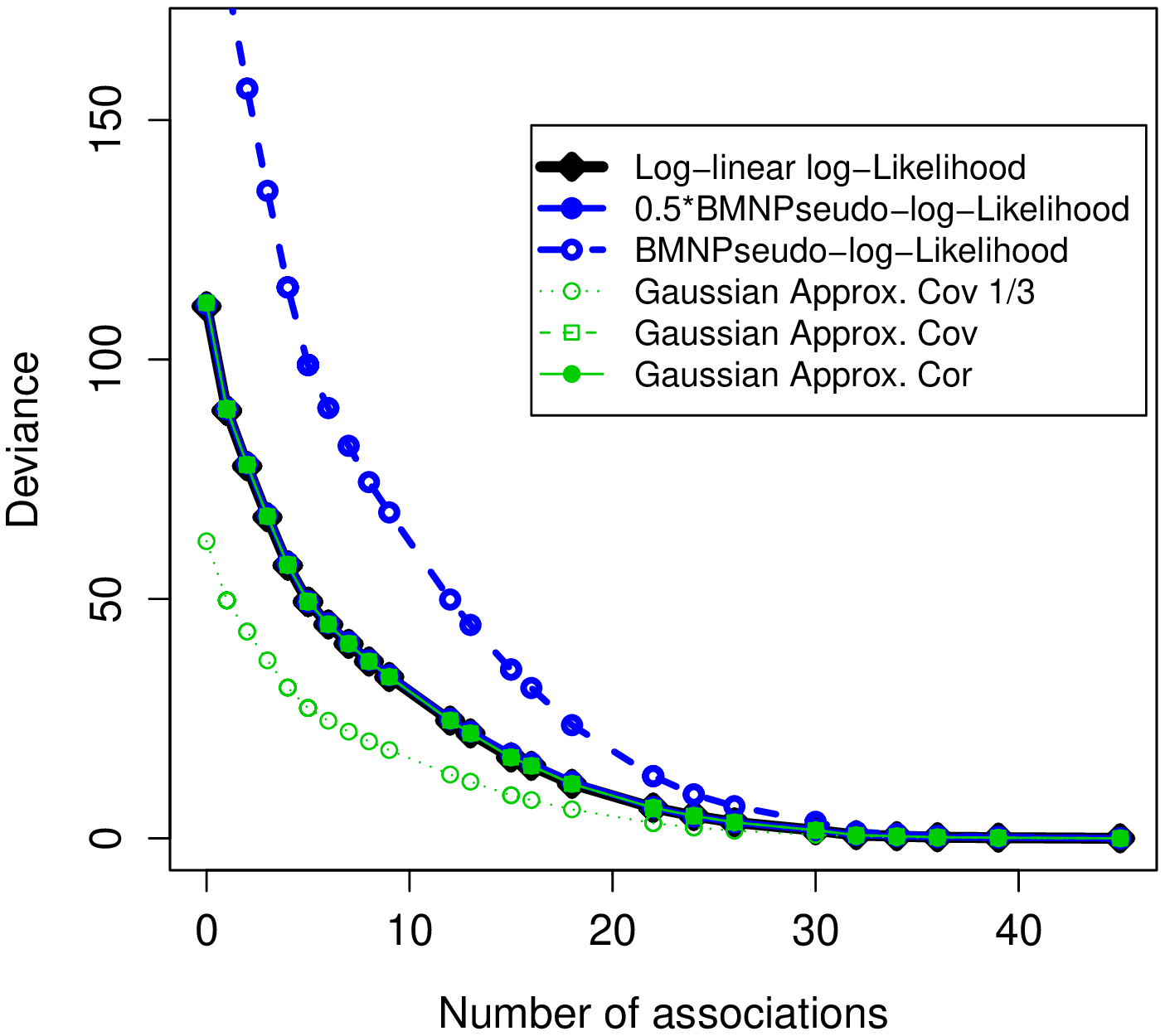}
\hspace{.15in}
\includegraphics[width=0.47\textwidth]{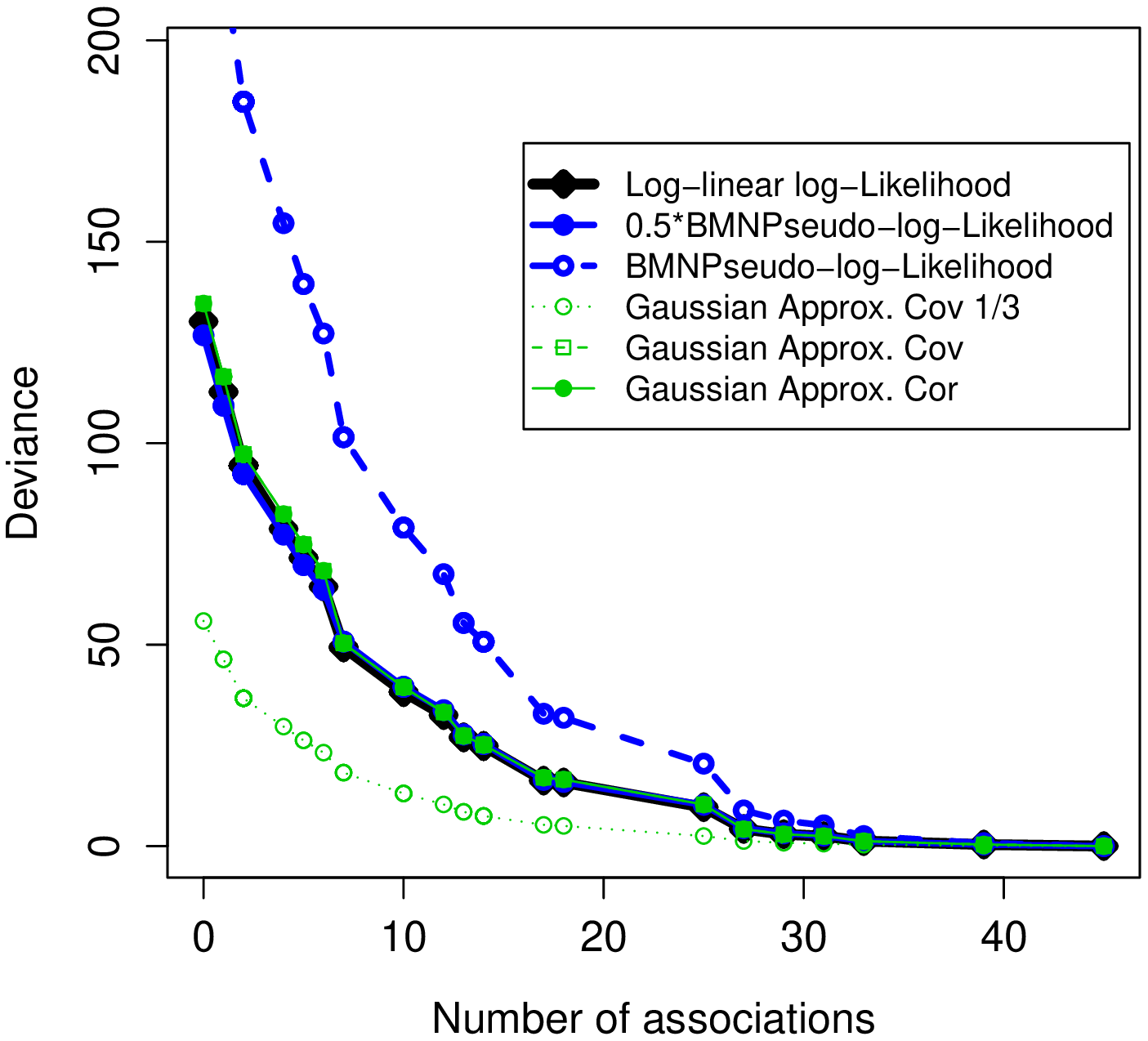}
\vspace{.15in}
\includegraphics[width=0.47\textwidth]{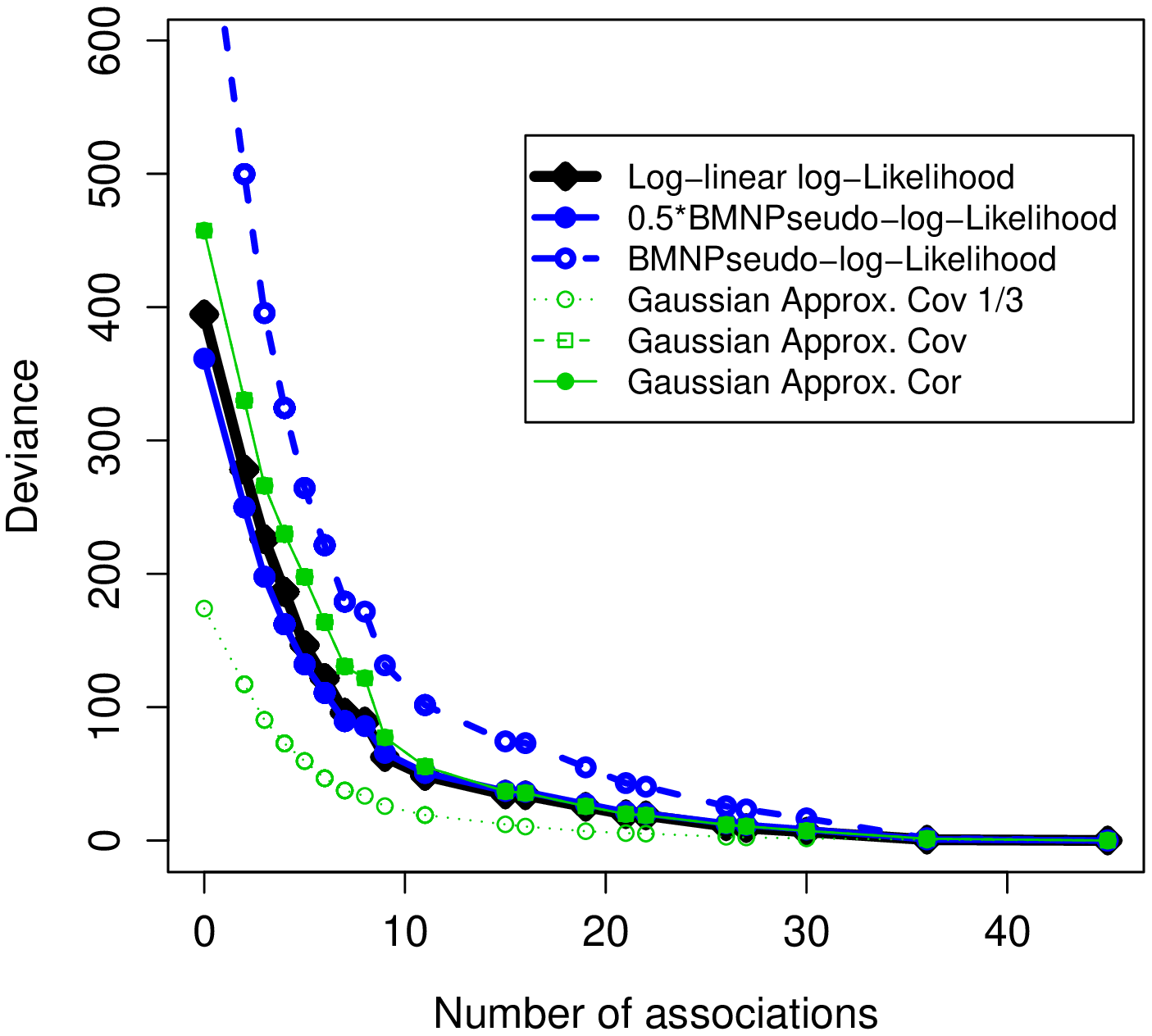}
\hspace{.15in}
\includegraphics[width=0.47\textwidth]{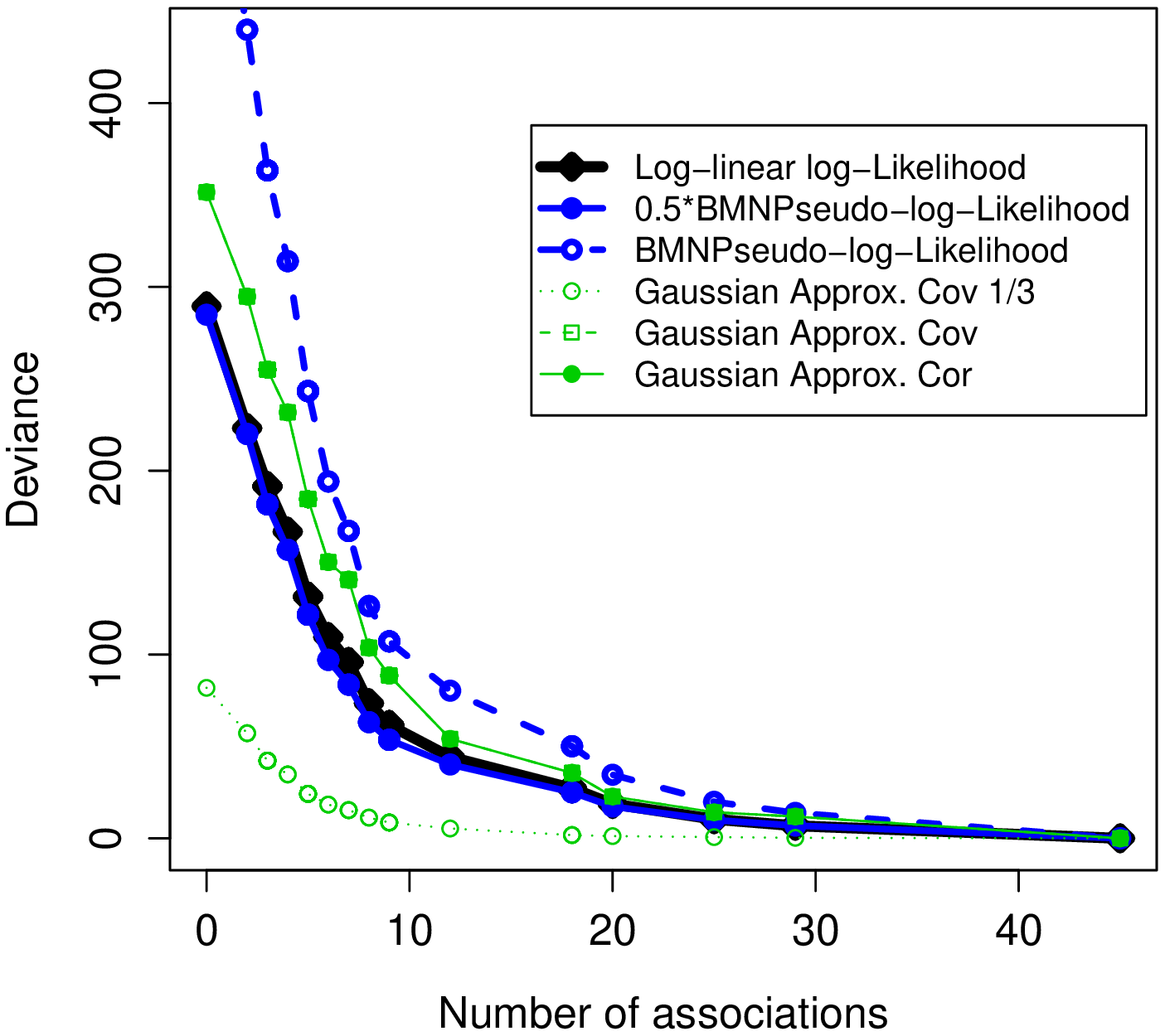}
\end{center}
\caption{Approximates for the Ising deviance. Deviances were
computed for each model selected by BMNPseudo for various values
of the sparsity parameter, on samples of size $n=500$ generated
with matrices $\Theta^{(1)}$ (upper left corner), $\Theta^{(2)}$
(upper right corner), $\Theta^{(3)}$ (lower left corner), and
$\Theta^{(4)}$ (lower right corner). Deviances were computed using
the exact log-linear log-likelihood (solid black line, solid
circles), the pseudo-likelihood (dashed blue line, circles), half
the pseudo-likelihood (solid blue line, solid circles), and the
Gaussian approximate log likelihoods based on the covariance
matrix with an additional 1/3 term on the diagonal (dotted green
line, circles), the covariance matrix (dashed green line, squares)
and the correlation matrix (solid green line, solid circles)(see
(\ref{pr_dev_pseudo})-(\ref{pr_dev_gauss}) for the corresponding
formula).}\label{Fig_Comp_Gauss_App_log_lik}
\end{figure}

\noindent Figure \ref{Fig_Comp_Gauss_App_log_lik} shows the
corresponding deviances. It can be seen that using the Gaussian
approximate log-likelihood based on the covariance matrix with the
additional 1/3 term on the diagonal results in a deviance which is
quite far from the exact one. Furthermore, the deviance obtained
with the covariance matrix (without adding the 1/3 term on the
diagonal) equals that obtained with the correlation matrix, and
both are closer to the exact deviance. Finally, the deviance of
the pseudo-(log)-likelihood is always greater than the exact
deviance. Using half the pseudo-likelihood corrects this
undesirable effect in most cases. \vskip5pt

\noindent These results should obviously be considered with
caution. Even if we tried to use various $\Theta$ matrices to
generate the data (and the conclusions were consistently the
same), a theoretical study would be needed to confirm these
empirical findings.

\subsection{Performance evaluation; the case $p=10$}
Let us first consider the performances achieved by the oracle
models (Tables \ref{table_simul1_Acc_1} and
\ref{table_simul1_Acc_2}). Overall, methods \verb"BMNPseudo" and
\verb"GaussCor" achieve good performances in terms of accuracy and
F1-score. It is also noteworthy that the computational time is
much higher for \verb"BMNPseudo", while overall performances of
\verb"GaussCor" appear to be slightly higher (especially under the
fourth simulation design). \vskip5pt

\noindent When focusing on the three \verb"GaussApprox" methods,
we observed the following ranking
$$\verb"GaussCor"\geq\verb"GaussCov"\geq\verb"GaussCov 1/3".$$
Consequently, and to save space, methods \verb"GaussCov"  and
\verb"GaussCov 1/3" will not be considered in the evaluation of
the models selected via the BIC procedure. It is still interesting
to note that \verb"GaussCor"$\geq$\verb"GaussCov" although we
observed that the approximate deviances under these two methods
were equal and close to the exact ones, which turn out to be an
insufficient condition for achieving good performances. \vskip5pt

\begin{table}
\begin{center}
\begin{footnotesize}
\caption{Evaluation of the "oracle models"; the case $p=10$. Means
(computed over 50 runs) are given for the computational time
needed to compute the models on a grid
of 50 equally-spaced $\lambda$ values as well as the number of
detected associations (POS), the false positive rate (FPR), the
true positive rate (TPR, which equals the recall, REC), the
precision (PRE), the accuracy (Acc.) and the F1 score
corresponding to the "oracle" model.} \label{table_simul1_Acc_1}
\begin{tabular}{l r@{.}l r@{.}l c c c c c}
\hline Method & \multicolumn{2}{c}{Time (s)}& \multicolumn{2}{c}{POS}& FPR& TPR$^\dag$& PRE& Acc.&F1 score\\
\hline
\multicolumn{10}{c}{\textbf{Data generated with $\Theta=\Theta^{(1)}$}}\\
\multicolumn{4}{l}{\emph{$n=100$}}\\
BMNPseudo& 6&47& 3&22& 0.034& 0.202& 0.724& 0.796& 0.291\\
GaussCov 1/3& 0&48& 1&84& 0.010& 0.150& 0.935& 0.804& 0.214\\
GaussCov& 0&48&2&02& 0.012& 0.160& 0.933& 0.804& 0.220\\
GaussCor& 0&48& 1&98& 0.011& 0.158& 0.937& 0.804& 0.219\\[0.05cm]
\multicolumn{4}{l}{\emph{$n=500$}}\\
BMNPseudo& 20&20& 6&96& 0.031& 0.588& 0.883& 0.884& 0.676\\
GaussCov 1/3& 0&72& 7&02& 0.033& 0.588& 0.877& 0.883& 0.674\\
GaussCov& 0&72& 7&00& 0.031& 0.590& 0.881& 0.884& 0.677\\
GaussCor& 0&72& 7&02& 0.031& 0.592& 0.881& 0.885& 0.678\\[0.05cm]
\multicolumn{4}{l}{\emph{$n=2500$}}\\
BMNPseudo& 82&65& 9&48& 0.003& 0.938& 0.991& 0.984& 0.961\\
GaussCov 1/3& 0&96& 9&36& 0.002& 0.928& 0.992& 0.982& 0.957\\
GaussCov& 0&96& 9&40& 0.002& 0.932& 0.992& 0.983& 0.959\\
GaussCor& 0&96& 9&40& 0.002& 0.932& 0.992& 0.983& 0.959\\
\\[0.2cm]
\multicolumn{10}{c}{\textbf{Data generated with $\Theta=\Theta^{(2)}$}}\\
\multicolumn{4}{l}{\emph{$n=100$}}\\
BMNPseudo& 20&17& 3&62& 0.024& 0.278& 0.864& 0.821& 0.376\\
GaussCov 1/3& 0&54& 3&72& 0.025& 0.284& 0.867& 0.821& 0.376\\
GaussCov& 0&54& 3&92& 0.028& 0.294& 0.857& 0.821& 0.386\\
GaussCor& 0&55& 3&60& 0.025& 0.274& 0.870& 0.820& 0.366
\\[0.05cm]
\multicolumn{4}{l}{\emph{$n=500$}}\\
BMNPseudo& 23&16& 7&02& 0.014& 0.654& 0.950& 0.912& 0.755\\
GaussCov 1/3& 0&79& 7&44& 0.02& 0.674& 0.927& 0.912& 0.761\\
GaussCov& 0&79& 7&46& 0.019& 0.678& 0.93& 0.913& 0.765\\
GaussCor& 0&86& 8&24& 0.017& 0.766& 0.938& 0.935& 0.832
\\[0.05cm]
\multicolumn{4}{l}{\emph{$n=2500$}}\\
BMNPseudo& 83&76& 9&74& 0.004& 0.960& 0.987& 0.988& 0.972\\
GaussCov 1/3& 0&96& 9&86& 0.008& 0.958& 0.975& 0.984& 0.964\\
GaussCov& 0&97& 9&70& 0.005& 0.954& 0.985& 0.986& 0.967\\
GaussCor& 0&99& 10&04& 0.002& 0.998& 0.995& 0.998& 0.996\\
\hline $^\dag$ TPR$=$REC.
\end{tabular}
\end{footnotesize}
\end{center}
\end{table}

\begin{table}
\begin{center}
\begin{footnotesize}
\caption{Evaluation of the "oracle models"; the case $p=10$. Means
(computed over 50 runs) are given for the computational time
needed to compute the models on a grid
of 50 equally-spaced $\lambda$ values as well as the number of
detected associations (POS), the false positive rate (FPR), the
true positive rate (TPR, which equals the recall, REC), the
precision (PRE), the accuracy (Acc.) and the F1 score
corresponding to the "oracle" model.} \label{table_simul1_Acc_2}
\begin{tabular}{l r@{.}l r@{.}l c c c c c}
\hline Method & \multicolumn{2}{c}{Time (s)}& \multicolumn{2}{c}{POS}& FPR& TPR$^\dag$& PRE& Acc.&F1 score\\
\hline
\multicolumn{10}{c}{\textbf{Data generated with $\Theta=\Theta^{(3)}$}}\\
\multicolumn{4}{l}{\emph{$n=100$}}\\
BMNPseudo& 35&08& 6&26& 0.018& 0.564& 0.929& 0.889& 0.678\\
GaussCov 1/3& 0&72& 6&54& 0.021& 0.582& 0.921& 0.891& 0.687\\
GaussCov& 0&72& 6&60& 0.022& 0.584& 0.917& 0.891& 0.689\\
GaussCor& 0&78& 7&66& 0.028& 0.668& 0.893& 0.904& 0.746
\\[0.05cm]
\multicolumn{4}{l}{\emph{$n=500$}}\\
BMNPseudo& 21&02& 8&76& 0.011& 0.838& 0.966& 0.956& 0.889\\
GaussCov 1/3& 0&84& 8&92& 0.027& 0.796& 0.913& 0.933& 0.836\\
GaussCov& 0&87& 8&88& 0.017& 0.828& 0.946& 0.948& 0.874\\
GaussCor& 0&96& 9&68& 0.006& 0.948& 0.981& 0.984& 0.963\\[0.05cm]
\multicolumn{4}{l}{\emph{$n=2500$}}\\
BMNPseudo& 80&60& 10&02& 0.002& 0.994& 0.993& 0.997& 0.993\\
GaussCov 1/3& 0&92& 10&84& 0.035& 0.962& 0.894& 0.964& 0.923\\
GaussCov& 0&97& 10&10& 0.007& 0.984& 0.976& 0.991& 0.979\\
GaussCor& 0&99& 10&00& 0.001& 0.998& 0.998& 0.999& 0.998\\
\\[0.2cm]
\multicolumn{10}{c}{\textbf{Data generated with $\Theta=\Theta^{(4)}$}}\\
\multicolumn{4}{l}{\emph{$n=100$}}\\
BMNPseudo& 39&89& 5&46& 0.055& 0.212& 0.816& 0.635& 0.312\\
GaussCov 1/3& 0&60& 6&84& 0.084& 0.245& 0.775& 0.633& 0.333\\
GaussCov& 0&60& 5&16& 0.053& 0.199& 0.800& 0.631& 0.297\\
GaussCor& 0&60& 6&86& 0.079& 0.253& 0.761& 0.639& 0.357
\\[0.05cm]
\multicolumn{4}{l}{\emph{$n=500$}}\\
BMNPseudo& 183&22& 12&94& 0.084& 0.566& 0.859& 0.768& 0.668\\
GaussCov 1/3& 0&68& 12&36& 0.08& 0.541& 0.853& 0.760& 0.651\\
GaussCov& 0&69& 13&02& 0.089& 0.563& 0.840& 0.764& 0.665\\
GaussCor& 0&75& 13&64& 0.055& 0.643& 0.905& 0.818& 0.745
\\[0.05cm]
\multicolumn{4}{l}{\emph{$n=2500$}}\\
BMNPseudo& 255&87& 15&06& 0.058& 0.714& 0.906& 0.846& 0.796\\
GaussCov 1/3& 0&76& 14&68& 0.072& 0.674& 0.880& 0.820& 0.759\\
GaussCov& 0&78& 15&16& 0.072& 0.700& 0.884& 0.832& 0.778\\
GaussCor& 0&85& 15&70& 0.037& 0.776& 0.944& 0.884& 0.848\\
\hline $^\dag$ TPR$=$REC.
\end{tabular}
\end{footnotesize}
\end{center}
\end{table}

\noindent Turning our attention to the evaluation of models
selected with the BIC procedure (Tables \ref{table_simul1_BIC_1}
and \ref{table_simul1_BIC_2}), a first observation is that, as
suggested by the results of Section
\ref{Sec_Comp_Approx_Deviances}, computing the BIC with half the
pseudo-likelihood (rather than the pseudo-likelihood itself)
results in better models in most cases. Moreover, from the
comparisons of the results of Tables \ref{table_simul1_Acc_1} and
\ref{table_simul1_Acc_2} and Tables \ref{table_simul1_BIC_1} and
\ref{table_simul1_BIC_2}, as $n$ grows, the BIC procedure appears
to enable to select models achieving performances similar to those
achieved by the "oracle" models. Moreover, the computation of the
un-shrunk estimates with method \verb"BMNPseudo" appears to be
very slow (the oracle models were much faster to compute than the
models selected by the BIC approach for this particular method,
especially when the sample size is small and in the fourth
simulation design). \vskip5pt

\noindent Overall, \verb"SepLogit OR", \verb"SepLogit AND" and
\verb"GaussCor"  are the best methods, closely followed by
\verb"BMNPseudo 1/2". Lastly, among these candidate methods,
\verb"GaussCor" is the fastest.\vskip5pt


\noindent We should lastly mention that method \verb"SepLogit" was
further tested using standardized covariates in each
$\ell_1$-penalized logistic regression models (results not shown).
To motivate this choice, we may mention that this is the default
option in package \verb"glmnet" as this approach is often adopted
in applications when using $\ell_1$-penalization (see
\cite{Boyd_logreg} for instance);  its suitability in our context
of binary variables was yet questionable. Interestingly, this
approach yielded results very similar to those obtained via the
"standard" one on data generated using matrices $\Theta^{(1)}$,
$\Theta^{(2)}$ and $\Theta^{(3)}$ and slightly better when using
matrix $\Theta^{(4)}$ (which however corresponds to the situation
where we observed the greatest variability in the performances of
every method).

\begin{table}
\begin{center}
\begin{footnotesize}
\caption{Evaluation of the models selected by the BIC procedure;
the case $p=10$. Means (computed over 50 runs) are given for the
computational time needed to compute the models on a grid
of 50 equally-spaced $\lambda$ values as well as for the number of
detected associations (POS), the false positive rate (FPR), the
true positive rate (TPR, which equals the recall, REC), the
precision (PRE), the accuracy (Acc.) and the F1 score
corresponding to the model selected by the BIC procedure.}
\label{table_simul1_BIC_1}
\begin{tabular}{l r@{.}l r@{.}l c c c c c}
\hline Method & \multicolumn{2}{c}{Time (s)}& \multicolumn{2}{c}{POS}& FPR& TPR$^\dag$& PRE& Acc.&F1 score\\
\hline
\multicolumn{10}{c}{\textbf{Data generated with $\Theta=\Theta^{(1)}$}}\\
\multicolumn{4}{l}{\emph{$n=100$}}\\
SepLogit OR& 21&94& 2&86& 0.043& 0.136& 0.540& 0.775& 0.234\\
SepLogit AND& 21&94& 2&12& 0.029& 0.112& 0.585& 0.780& 0.216\\
BMNPseudo & 14&20& 8&16& 0.141& 0.322& 0.424& 0.740& 0.355\\
BMNPseudo 1/2 & 14&20& 2&60& 0.039& 0.122& 0.492& 0.774& 0.233\\
GaussCor & 1&12& 2&56& 0.035& 0.132& 0.572& 0.780& 0.235
\\[0.1cm]
\multicolumn{4}{l}{\emph{$n=500$}}\\
SepLogit OR& 22&91& 4&80& 0.018& 0.416& 0.885& 0.856& 0.549\\
SepLogit AND& 22&91& 4&12& 0.014& 0.364& 0.892& 0.848& 0.504\\
BMNPseudo & 42&43& 9&48& 0.079& 0.670& 0.725& 0.865& 0.687\\
BMNPseudo 1/2 & 42&43& 4&60& 0.015& 0.408& 0.895& 0.857& 0.550\\
GaussCor & 1&16& 4&48& 0.014& 0.400& 0.895& 0.856& 0.539
\\[0.1cm]
\multicolumn{4}{l}{\emph{$n=2500$}}\\
SepLogit OR& 27&89& 9&52& 0.006& 0.93& 0.979& 0.980& 0.952\\
SepLogit AND& 27&89& 9&32& 0.004& 0.918& 0.986& 0.979& 0.949\\
BMNPseudo & 176&09& 11&66& 0.055& 0.972& 0.846& 0.951& 0.901\\
BMNPseudo 1/2 & 176&09& 9&36& 0.005& 0.920& 0.985& 0.979& 0.948\\
GaussCor & 1&19& 9&30& 0.005& 0.912& 0.983& 0.976& 0.943\\
\\[0.2cm]
\multicolumn{10}{c}{\textbf{Data generated with $\Theta=\Theta^{(2)}$}}\\
\multicolumn{4}{l}{\emph{$n=100$}}\\
SepLogit OR& 23&45& 4&68& 0.058& 0.266& 0.618& 0.792& 0.361\\
SepLogit AND& 23&45& 2&76& 0.027& 0.182& 0.687& 0.797& 0.302\\
BMNPseudo & 215&65& 9&82& 0.154& 0.444& 0.482& 0.757& 0.443\\
BMNPseudo 1/2 & 215&65& 3&22& 0.033& 0.208& 0.688& 0.799& 0.325\\
GaussCor & 1&11& 3&80& 0.041& 0.236& 0.686& 0.798& 0.341
\\[0.1cm]
\multicolumn{4}{l}{\emph{$n=500$}}\\
SepLogit OR& 33&18& 7&20& 0.018& 0.658& 0.922& 0.910& 0.758\\
SepLogit AND& 33&18& 6&18& 0.007& 0.592& 0.960& 0.904& 0.720\\
BMNPseudo & 97&70& 10&82& 0.087& 0.776& 0.735& 0.882& 0.747\\
BMNPseudo 1/2 & 97&70& 6&56& 0.014& 0.606& 0.934& 0.901& 0.720\\
GaussCor & 1&16& 7&02& 0.014& 0.652& 0.931& 0.912& 0.758
\\[0.1cm]
\multicolumn{4}{l}{\emph{$n=2500$}}\\
SepLogit OR& 34&52& 10&28& 0.011& 0.990& 0.966& 0.989& 0.977\\
SepLogit AND& 34&52& 9&92& 0.005& 0.976& 0.985& 0.991& 0.980\\
BMNPseudo & 175&80& 11&70& 0.049& 0.998& 0.872& 0.961& 0.926\\
BMNPseudo 1/2 & 175&80& 10&16& 0.013& 0.972& 0.960& 0.984& 0.965\\
GaussCor & 1&30& 10&06& 0.005& 0.988& 0.984& 0.993& 0.985\\
\hline $^\dag$ TPR$=$REC.
\end{tabular}
\end{footnotesize}
\end{center}
\end{table}

\begin{table}
\begin{center}
\begin{footnotesize}
\caption{Evaluation of the models selected by the BIC procedure;
the case $p=10$. Means (computed over 50 runs) are given for the
computational time needed to compute the models on a grid
of 50 equally-spaced $\lambda$ values as well as the number of
detected associations (POS), the false positive rate (FPR), the
true positive rate (TPR, which equals the recall, REC), the
precision (PRE), the accuracy (Acc.) and the F1 score
corresponding to the model selected by the BIC procedure.}
\label{table_simul1_BIC_2}
\begin{tabular}{l r@{.}l r@{.}l c c c c c}
\hline Method & \multicolumn{2}{c}{Time (s)}& \multicolumn{2}{c}{POS}& FPR& TPR$^\dag$& PRE& Acc.&F1 score\\
\hline
\multicolumn{10}{c}{\textbf{Data generated with $\Theta=\Theta^{(3)}$}}\\
\multicolumn{4}{l}{\emph{$n=100$}}\\
SepLogit OR& 30&86& 7&86& 0.055& 0.594& 0.775& 0.867& 0.660\\
SepLogit AND& 30&86& 5&44& 0.021& 0.470& 0.868& 0.866& 0.599\\
BMNPseudo & 276&48& 12&18& 0.147& 0.702& 0.613& 0.819& 0.640\\
BMNPseudo 1/2 & 276&48& 6&44& 0.031& 0.534& 0.856& 0.872& 0.638\\
GaussCor & 1&26& 7&42& 0.037& 0.612& 0.834& 0.885& 0.697
\\[0.1cm]
\multicolumn{4}{l}{\emph{$n=500$}}\\
SepLogit OR& 33&10& 10&14& 0.027& 0.918& 0.912& 0.960& 0.912\\
SepLogit AND& 33&10& 8&90& 0.009& 0.858& 0.967& 0.961& 0.906\\
BMNPseudo & 163&20& 12&78& 0.098& 0.936& 0.760& 0.910& 0.830\\
BMNPseudo 1/2 & 163&20& 9&80& 0.030& 0.876& 0.906& 0.949& 0.885\\
GaussCor & 1&26& 9&78& 0.014& 0.930& 0.957& 0.974& 0.940
\\[0.1cm]
\multicolumn{4}{l}{\emph{$n=2500$}}\\
SepLogit OR& 40&76& 10&30& 0.009& 1.000& 0.973& 0.993& 0.986\\
SepLogit AND& 40&76& 10&02& 0.001& 1.000& 0.998& 0.999& 0.999\\
BMNPseudo & 176&22& 11&14& 0.033& 1.000& 0.914& 0.975& 0.951\\
BMNPseudo 1/2 & 176&22& 10&18& 0.005& 1.000& 0.984& 0.996& 0.992\\
GaussCor & 1&26& 10&12& 0.003& 1.000& 0.989& 0.997& 0.994\\
\\[0.2cm]
\multicolumn{10}{c}{\textbf{Data generated with $\Theta=\Theta^{(4)}$}}\\
\multicolumn{4}{l}{\emph{$n=100$}}\\
SepLogit OR& 30&63& 20&96& 0.443& 0.497& 0.452& 0.532& 0.470\\
SepLogit AND& 30&63& 14&88& 0.322& 0.343& 0.440& 0.537& 0.383\\
BMNPseudo & 193&43& 27&16& 0.577& 0.640& 0.446& 0.515& 0.523\\
BMNPseudo 1/2 & 193&43& 18&70& 0.405& 0.431& 0.440& 0.526& 0.431\\
GaussCor & 1&87& 20&72& 0.452& 0.473& 0.432& 0.516& 0.445
\\[0.1cm]
\multicolumn{4}{l}{\emph{$n=500$}}\\
SepLogit OR& 36&10& 10&48& 0.038& 0.500& 0.916& 0.767& 0.642\\
SepLogit AND& 36&10& 8&30& 0.012& 0.421& 0.967& 0.749& 0.583\\
BMNPseudo & 1090&87& 13&52& 0.107& 0.565& 0.815& 0.755& 0.657\\
BMNPseudo 1/2 & 1090&87& 9&84& 0.044& 0.458& 0.897& 0.746& 0.600\\
GaussCor & 1&80& 10&70& 0.025& 0.529& 0.945& 0.787& 0.675
\\[0.1cm]
\multicolumn{4}{l}{\emph{$n=2500$}}\\
SepLogit OR& 52&32& 15&26& 0.048& 0.737& 0.922& 0.861& 0.817\\
SepLogit AND& 52&32& 13&44& 0.016& 0.685& 0.971& 0.858& 0.802\\
BMNPseudo& 2299&82& 17&40& 0.122& 0.748& 0.829& 0.823& 0.783\\
BMNPseudo 1/2 & 2299&82& 14&72& 0.056& 0.698& 0.905& 0.840& 0.786\\
GaussCor & 1&86& 14&76& 0.026& 0.741& 0.957& 0.876& 0.834\\
\hline $^\dag$ TPR$=$REC.
\end{tabular}
\end{footnotesize}
\end{center}
\end{table}

\subsection{Performance evaluation; the case $p=50$}
To save space, we only present here the performances achieved by
models selected via the BIC procedure, on samples of size $n=500$
and $n=2500$ generated using either matrix
diag$(\Theta^{(j)},\Theta^{(j)},\Theta^{(j)},\Theta^{(j)},\Theta^{(j)})$,
for $j=1,2,3$ or matrix $\Theta^{(5)}$. Moreover, the results
obtained in the case $p=10$ especially show that method
\verb"BMNPseudo" can be quite slow, and that it does not
outperform method \verb"SepLogit". Lastly, among the methods
relying on a Gaussian approximate of the Ising likelihood, method
\verb"GaussCor" was observed to be the best. Therefore, in order
to save computational time, only methods \verb"GaussCor" and
\verb"SepLogit" were considered in the case $p=50$. \vskip5pt

\noindent Results are presented in Table \ref{table_simul1_BIC50}.
They are consistent with what was observed in the case $p=10$.
More precisely, methods \verb"SepLogit" and \verb"GaussCor"
achieve comparable performances. Regarding computational time,
\verb"GaussCor" is still significantly faster than
\verb"SepLogit".

\begin{table}
\begin{center}
\begin{footnotesize}
\caption{Evaluation of the models selected by the BIC procedure;
the case $p=50$. Means (computed over 50 runs) are given for the
computational time needed to compute the models on a grid
of 50 equally-spaced $\lambda$ values as well as the number of
detected associations (POS), the false positive rate (FPR), the
true positive rate (TPR, which equals the recall, REC), the
precision (PRE), the accuracy (Acc.) and the F1 score
corresponding to the model selected by the BIC procedure.}
\label{table_simul1_BIC50}
\begin{tabular}{l r@{.}l r@{.}l c c c c c}
\hline Method & \multicolumn{2}{c}{Time (s)}& \multicolumn{2}{c}{POS}& FPR& TPR$^\dag$& PRE& Acc.&F1 score\\
\hline
\multicolumn{10}{c}{\textbf{Data generated with $\Theta=\mbox{diag}(\Theta^{(1)},\Theta^{(1)},\Theta^{(1)},\Theta^{(1)},\Theta^{(1)})$}}\\
\multicolumn{4}{l}{\emph{$n=500$}}\\
SepLogit OR&  236&21& 39&00& 0.016& 0.399& 0.515& 0.960&0.447 \\
SepLogit AND& 236&21& 30&10& 0.011& 0.352& 0.588& 0.963& 0.438\\
GaussCor&      31&85& 34&06& 0.013& 0.376& 0.562& 0.962& 0.446
\\[0.05cm]
\multicolumn{4}{l}{\emph{$n=2500$}}\\
SepLogit OR&  248&74& 54&48& 0.006& 0.938& 0.864& 0.991& 0.898\\
SepLogit AND& 248&74& 50&64& 0.004& 0.921& 0.910& 0.993& 0.915\\
GaussCor     & 28&86& 52&34& 0.005& 0.918& 0.881& 0.991& 0.898\\
\\[0.2cm]
\multicolumn{10}{c}{\textbf{Data generated with $\Theta=\mbox{diag}(\Theta^{(2)},\Theta^{(2)},\Theta^{(2)},\Theta^{(2)},\Theta^{(2)})$}}\\
\multicolumn{4}{l}{\emph{$n=500$}}\\
SepLogit OR&  285&91& 53&36& 0.017& 0.661& 0.626& 0.970& 0.640\\
SepLogit AND& 285&91& 34&14& 0.007& 0.527& 0.778& 0.974& 0.626\\
GaussCor      & 32&75& 45&42& 0.011& 0.642& 0.717& 0.975& 0.673
\\[0.05cm]
\multicolumn{4}{l}{\emph{$n=2500$}}\\
SepLogit OR&  304&09& 59&26& 0.008& 0.988& 0.837& 0.991& 0.905\\
SepLogit AND& 304&09& 50&80& 0.003& 0.950& 0.937& 0.995& 0.943\\
GaussCor     & 29&00& 54&40& 0.004& 0.995& 0.917& 0.996& 0.954\\
\\[0.2cm]
\multicolumn{10}{c}{\textbf{Data generated with $\Theta=\mbox{diag}(\Theta^{(3)},\Theta^{(3)},\Theta^{(3)},\Theta^{(3)},\Theta^{(3)})$}}\\
\multicolumn{4}{l}{\emph{$n=500$}}\\
SepLogit OR&  305&44& 68&66& 0.019& 0.936& 0.685& 0.980& 0.790\\
SepLogit AND& 305&44& 47&04& 0.005& 0.834& 0.888& 0.989& 0.859\\
GaussCor     & 30&34& 60&34& 0.010& 0.968& 0.811& 0.989& 0.880\\[0.05cm]
\multicolumn{4}{l}{\emph{$n=2500$}}\\
SepLogit OR&  380&86& 57&72& 0.007& 1.000& 0.868& 0.994& 0.929\\
SepLogit AND& 380&86& 51&50& 0.001& 1.000& 0.971& 0.999& 0.985\\
GaussCor     & 25&48& 50&48& 0.000& 1.000& 0.991& 1.000& 0.995\\
\\[0.2cm]
\multicolumn{10}{c}{\textbf{Data generated with $\Theta=\Theta^{(5)}$}}\\
\multicolumn{4}{l}{\emph{$n=500$}}\\
SepLogit OR& 271&84& 95&20& 0.017& 0.610& 0.802& 0.945& 0.692\\
SepLogit AND& 271&84& 71&00& 0.007& 0.508& 0.894& 0.944& 0.647\\
GaussCor    & 74&71& 86&96& 0.013& 0.582& 0.839& 0.946& 0.685\\[0.05cm]
\multicolumn{4}{l}{\emph{$n=2500$}}\\
SepLogit OR& 307&38& 129&00& 0.008& 0.960& 0.931& 0.989& 0.945\\
SepLogit AND& 307&38& 117&48& 0.002& 0.922& 0.981& 0.990& 0.950\\
GaussCor     & 65&41& 123&68& 0.007& 0.927& 0.938& 0.986& 0.932\\
\hline $^\dag$ TPR$=$REC.
\end{tabular}
\end{footnotesize}
\end{center}
\end{table}

\subsection{Agreement between the compared methods} One way to
measure the agreement between two selected models is to compare
them with their intersection. Let $\mathcal{G}_1=({V},{E}_1)$ and
$\mathcal{G}_2=({V},{E}_2)$ be the two graphs to be compared.
Further let ${E}^\star={E}_1\cap{E}_2$ and denote by $\sharp {E}$
the cardinality of a set $E$ of edges. To compare graphs
$\mathcal{G}_1$ and $\mathcal{G}_2$, we will consider the
quantities
\begin{eqnarray}
\kappa(\mathcal{G}_1,\mathcal{G}_2)&=&\frac{\sharp
{E}^\star}{\min(\sharp {E}_1,\sharp {E}_2)}, \label{agree}\\
\bar\kappa(\mathcal{G}_1,\mathcal{G}_2)&=&(\sharp {E}_1-\sharp
{E}^\star) +(\sharp {E}_2-\sharp {E}^\star), \label{disagree}
\end{eqnarray}
as measures of agreement and disagreement respectively
($\bar\kappa$ is the cardinality of the symmetric difference
between ${E}_1$ and ${E}_2$). Observe that according to these
measures, the saturated graph would both agree and disagree "a
lot" with any sub-graph. \vskip5pt

\noindent Results are presented in Table \ref{table_Agree} for
$p=10$ and $\Theta=\{\Theta^{(3)},\Theta^{(4)}\}$ which correspond
to the two extreme situations in terms of signal-to-noise ratio.
Overall, $\verb"BMNPseudo"$ is closer to $\verb"SepLogit"$ than
$\verb"GaussCor"$, what was expected given the respective
principles of the methods. Moreover, agreement [resp.
disagreement] between the various models is higher [resp. lower]
when the signal-to-noise ratio is high, that is when $n=2500$
and/or $\Theta=\Theta^{(3)}$. When the signal-to-noise ratio is
low and models are quite different, a natural question is how to
get the best model. Intersecting two models is one of the
candidate approaches. For the sake of brevity, we do not present
the complete results here, but intersecting \verb"GaussCor" and
\verb"SepLogit OR" for instance resulted in quite conservative
models that generally achieved better performances than either
\verb"GaussCor" or \verb"SepLogit OR" (in terms of accuracy and
F1-score). \vskip5pt

\noindent Models derived under method \verb"SepLogit" with
standardized covariates were also compared to the other models
(results not shown). Overall, we observed very good agreement
between the standard approach and the standardized approach
(especially in the case of high signal-to-noise ratio). We also
observed slightly better agreements between these models and those
derived under method \verb"GaussCor", especially on datasets
generated using matrix $\Theta^{(4)}$.

\begin{table}
\begin{center}
\begin{footnotesize}
\caption{Results of the simulation study: agreement evaluation.
Means (along with standard deviations) of the criteria defined in
(\ref{agree})-(\ref{disagree}) were obtained from 50 runs for
various sample sizes $n$ and matrices
$\Theta$.}\label{table_Agree}
\begin{tabular}{l c c c c }
\hline Comparisons  & \multicolumn{2}{c}{Agreement $\kappa$} &
\multicolumn{2}{c}{Disagreement $\bar \kappa$}  \\
\hline
\multicolumn{5}{l}{\textbf{Data generated with $\Theta=\Theta^{(3)}$}, $n=100$}\\
SepLogit OR SepLogit AND& 1.000&   (0.000)&   2.420&    (1.605)   \\
SepLogit OR BMNPseudo&    0.970&   (0.071)&   2.140&    (1.807)   \\
SepLogit AND BMNPseudo&   0.965&   (0.099)&   1.720&    (1.578)   \\
BMNPseudo GaussCor&       0.941&   (0.087)&   2.940&    (1.463)   \\
SepLogit OR GaussCor&     0.941&   (0.077)&   2.480&    (1.632)   \\
SepLogit AND GaussCor&    0.940&   (0.086)&   3.060&    (1.476)
\\[0.2cm]
\multicolumn{5}{l}{\textbf{Data generated with $\Theta=\Theta^{(3)}$}, n=2500}\\
SepLogit OR SepLogit AND& 1.000&   (0.000)&   0.280&    (0.497)   \\
SepLogit OR BMNPseudo&    1.000&   (0.000)&   0.160&    (0.370)   \\
SepLogit AND BMNPseudo&   0.998&   (0.013)&   0.200&    (0.495)   \\
BMNPseudo GaussCor&       1.000&   (0.000)&   0.100&    (0.303)   \\
SepLogit OR GaussCor&     1.000&   (0.000)&   0.220&    (0.418)   \\
SepLogit AND GaussCor&    0.998&   (0.013)&   0.140&    (0.405)
\\[0.5cm]
\multicolumn{5}{l}{\textbf{Data generated with $\Theta=\Theta^{(4)}$}, $n=100$}\\
SepLogit OR SepLogit AND& 1.000&   (0.000)&   6.080&     (2.146)   \\
SepLogit OR BMNPseudo&    0.952&   (0.048)&   4.800&     (2.711)   \\
SepLogit AND BMNPseudo&   0.957&   (0.053)&   5.400&     (2.148)   \\
BMNPseudo GaussCor&       0.967&   (0.054)&   6.740&     (2.863)   \\
SepLogit OR GaussCor&     0.896&   (0.081)&   8.380&     (2.900)   \\
SepLogit AND GaussCor&    0.931&   (0.081)&   9.900&     (3.112)
\\[0.2cm]
\multicolumn{5}{l}{\textbf{Data generated with $\Theta=\Theta^{(4)}$}, n=2500}\\
SepLogit OR SepLogit AND& 1.000&   (0.000)&   1.820&    (1.424)   \\
SepLogit OR BMNPseudo&    0.981&   (0.031)&   1.880&    (1.335)   \\
SepLogit AND BMNPseudo&   0.983&   (0.030)&   1.900&    (1.359)   \\
BMNPseudo GaussCor&       0.950&   (0.050)&   2.900&    (1.474)   \\
SepLogit OR GaussCor&     0.972&   (0.035)&   2.220&    (1.183)   \\
SepLogit AND GaussCor&    0.980&   (0.035)&   2.200&    (1.666)   \\
\hline
\end{tabular}
\end{footnotesize}
\end{center}
\end{table}

\subsection{Comparison of the conditional odds-ratios estimates}
Both methods \verb"SepLogit" and \verb"BMNPseudo" have been
empirically shown to yield appropriate estimates for the
conditional odds-ratios. On the other hand, it is rather unclear
whether estimates derived from methods relying on the Gaussian
approximation would be consistent. We therefore conclude this
simulation study with a simple study to check it. \vskip5pt

\noindent To avoid interpretation difficulties related to the
model selection accuracy, coefficient estimates were computed
under the sparsity structure of the true models and compared with
the true coefficients (this corresponds to the situation where the
true sparsity structure is known). To do so, we used an approach
similar to the one used to get un-shrunk estimates for the BIC
procedure. \vskip5pt

\noindent The mean squared errors for the conditional log-odds
ratios, which we defined here as
\begin{equation}\label{MSE}
\mbox{MSE}=1000 \times \frac{\sum_{k> \ell} (\hat\theta_{k,\ell}-
\theta_{k,\ell})^2}{\sum_{k> \ell}\1\{\theta_{k,\ell}\neq 0\}},
\end{equation}
are reported in Table \ref{table_SCE} for methods \verb"SepLogit",
\verb"GaussCov 1/3" and \verb"GaussCor" in the case $p=10$ and for
samples of sizes $n=500$ and $n=2500$ (for \verb"SepLogit" each
coefficient $\hat\theta_{k,\ell}$ was set as the mean of the two
coefficients returned by the two constrained logistic regressions
involved in this method). It can be observed that overall neither
\verb"GaussCov 1/3" nor \verb"GaussCor" led to consistent
estimates for the $\theta_{k,\ell}$ coefficients. These methods
(especially \verb"GaussCor") should therefore be combined with
other methods (for instance \verb"SepLogit") when estimates of the
conditional odds-ratios are needed. \vskip5pt

\noindent Inconsistency of the estimates derived under method
\verb"GaussCor" can be explained as follows. Since this method
relies on the correlation matrix, it is closely related to the
method consisting in performing linear regressions at each node
(as shown by \cite{MeinBuhl} in the Gaussian case). Therefore, the
coefficients returned by this method are related to the
coefficients $\gamma_{k,\ell}$ involved in the linear model
\begin{equation}
\E(X^{(k)}|\bx^{(-k)})=\P(X^{(k)}=1|\bx^{(-k)})=\ \gamma_{k}+
\sum_{\ell\neq k}\gamma_{k,\ell}x^{(\ell)}.\nonumber
\end{equation}
Clearly, coefficients $\gamma_{k,\ell}$ are quite different from
the conditional log odds-ratios $\theta_{k,\ell}$ involved in the
Ising model (see (\ref{Ising})).

\begin{table}
\begin{center}
\begin{footnotesize}
\caption{Results of the simulation study: evaluation of the
conditional log-odds-ratios estimation. Means of the mean squared
error (MSE) defined in (\ref{MSE}) were obtained from 50 runs for
various sample sizes $n$ and matrices $\Theta$. Three methods were
evaluated: the genuine method of Banerjee et al. (2008) GaussCov
1/3, its modification relying on the correlation matrix GaussCor
and the method relying on multiple logistic regressions
SepLogit.}\label{table_SCE}
\begin{tabular}{l r@{.}l r@{.}l r@{.}l }
\hline Sample size  & \multicolumn{2}{c}{GaussCov 1/3} &
\multicolumn{2}{c}{GaussCor} &
\multicolumn{2}{c}{SepLogit} \\
\hline
\multicolumn{5}{l}{\textbf{Data generated with $\Theta=\Theta^{(1)}$}}\\
$n=500$&  2&613&   2&073&   1&957\\
$n=2500$& 2&140&   0&451&   0&435\\[0.3cm]
\multicolumn{5}{l}{\textbf{Data generated with $\Theta=\Theta^{(2)}$}}\\
$n=500$&  14&568&  7&743&  6&175\\
$n=2500$& 14&342&  6&573&  1&201\\[0.3cm]
\multicolumn{5}{l}{\textbf{Data generated with $\Theta=\Theta^{(3)}$}}\\
$n=500$&  58&502&  29&197&  10&704\\
$n=2500$& 57&451&  26&779&  1&739\\[0.3cm]
\multicolumn{5}{l}{\textbf{Data generated with $\Theta=\Theta^{(4)}$}}\\
$n=500$&  98&304&  68&034&  50&007\\
$n=2500$& 97&402&  65&973&  12&794\\
\hline
\end{tabular}
\end{footnotesize}
\end{center}
\end{table}

\section{Application to the search for associations among causes of death}\label{Sec_Results}

The general objective of the application in this example is to
detect associations between  causes of death and identify possibly
relevant groupings of causes contributing to the death.

\subsection{Description of the data}
The dataset we used consists in causes of death recorded in all
death certificates ($n=535\,684$) in France for the year 2005. In
France, death certification (compulsory with 100\% coverage)
conforms to the WHO guideline: the death process is described
starting from the underlying causes of death and ending with the
immediate cause of death; other contributing causes of death are
also recorded. All these causes were considered in the analysis.
Causes are further coded according to the International
Classification of Diseases (ICD)(in 2005, the tenth revision
\citep{ICD10}). The total number of code categories is large
(about 4000 codes used) but for this analysis we applied a
simplified classification of 59 entities according to the Eurostat
shortlist \citep{Shortlist} (see Appendix A for the classification
used). Therefore, $p=59$ and each death certificate can be
regarded as a vector $\bx=(x^{(1)},...,x^{(59)})$ in
$\{0,1\}^{59}$, where, for all $k=1,...,59$, $x^{(k)}=1$ if and
only if the $k$-th class is recorded on the death certificate.
About 11\%  of the certificates had more than five causes of
death, 14\% had four causes, 26\% had three, 30\% had two and 18\%
had a sole cause of death. The frequencies of each cause are
reported in Appendix A; the most frequent causes of death were, in
decreasing order, heart failure, ischaemic heart diseases,
cerebrovascular diseases, hypertensive diseases, pneumonia,
diabetes, lung cancer and senility.

\subsection{Graphical model analysis}\label{Res_choice_sparsity}

\noindent Most common causes of death clearly depend upon age and
gender, and a natural question is whether associations among
causes of death also vary with age and gender. We then decided to
split the whole population into strata defined by gender and age
The complete analysis of every stratum is out of the scope of the
present paper, and we only present here the results from the
analysis of two sub-groups, namely those of males aged between 15
and 24 (2918 observations) and males aged between 45 and 64 (57045
observations). \vskip5pt

\noindent First considering the group of age 15-24, we applied
\verb"GaussCor", \verb"BMNPseudo", \verb"SepLogit AND", and
\verb"SepLogit OR" after selecting the sparsity parameter
according to the BIC procedure described above. This yielded
models with 113, 107, 61 and 129 associations respectively. Good
agreement was observed between models derived under methods
\verb"SepLogit" and \verb"BMNPseudo" (we had $\kappa=0.935$ for
the comparison between \verb"BMN" and \verb"SepLogit OR" for
instance). However, the model derived under method \verb"GaussCor"
was slightly different from the three other models (we had
$\kappa=0.700$ for the comparison between \verb"GaussCor" and
\verb"SepLogit OR" and $\kappa=0.918$ for the comparison between
\verb"GaussCor" and \verb"SepLogit AND"). More precisely, the
model obtained with method \verb"GaussCor" entailed many more
positive associations, a few of which corresponding to variables
co-occurring only once in the sub-group. This suggests that method
\verb"GaussCor" might be a little too sensitive to positive
associations, especially when the signal-to-noise ratio is low (in
this particular study, the signal-to-noise ratio is low due to
highly unbalanced variables). Regarding computational time, 1.6
second was needed to compute method \verb"GaussCor" while it took
19628 and 876 seconds for computing methods \verb"BMNPseudo" and
\verb"SepLogit" respectively (analyses were performed on the
Windows machine). For these latter two methods, we were not able
to conduct the analysis with the choice
$\lambda^{\footnotesize{\mbox{min}}}=\lambda^{\footnotesize{\mbox{max}}}/1000$,
and we had to select
$\lambda^{\footnotesize{\mbox{min}}}=\lambda^{\footnotesize{\mbox{max}}}/50$
and
$\lambda^{\footnotesize{\mbox{min}}}=\lambda^{\footnotesize{\mbox{max}}}/100$
for methods \verb"BMNPseudo" and \verb"SepLogit" respectively. It
is also noteworthy that the computational time needed for methods
\verb"BMNPseudo" and \verb"SepLogit" is mostly due to the
computation of the un-shrunk estimates (necessary for the
derivation of the BIC); omitting this step, the computational time
using method \verb"BMNPseudo" [resp. \verb"SepLogit"] is reduced
to 4598 seconds [resp. 198 seconds]. \vskip5pt

\begin{figure}
\begin{center}
\includegraphics[width=0.90\textwidth]{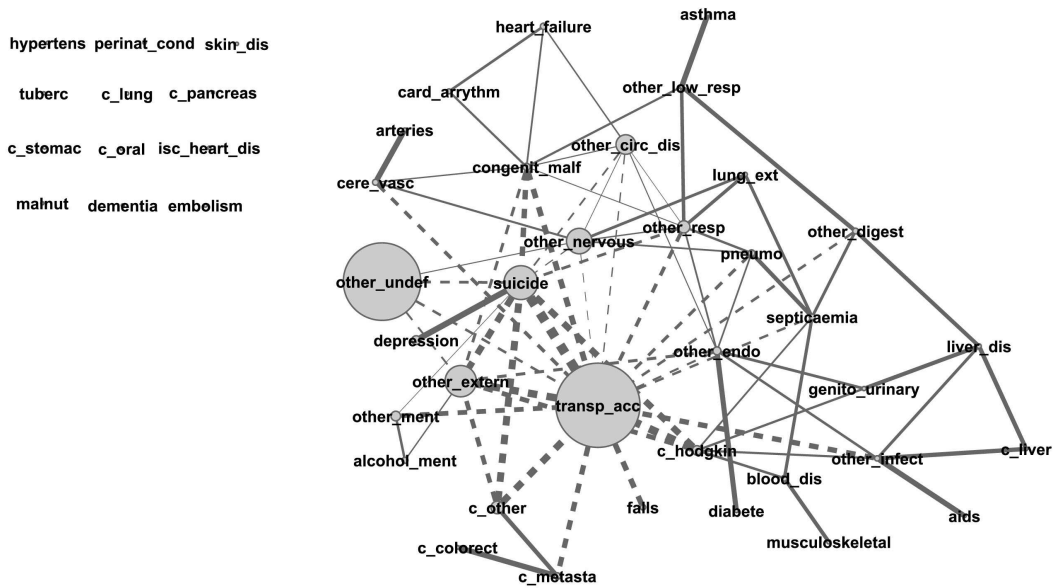}
\caption{Graphical model obtained with Cytoscape on the real data
set (males aged between 15 and 24 years). Positive associations
(solid lines) and negative associations (dashed lines) are
presented. The line widths of edges are proportional to the
conditional log-odds-ratios (estimated using multiple logistic
regressions built under the constraint implied by the retained
model). For instance, the (absolute value of the) conditional
log-odds-ratio was about 4.5 for the association
depression/suicide, 3 for the association liver disease/other
diseases of the digestive system, 1.7 for the association other
infectious disease/other endocrinal disease, and 0.35 for the
association other mental disorderd/suicide. Similarly, vertices
are represented by balls with diameter related to the observed
frequency of the causes of death. Transport accidents were
reported on about 40\% of the death certificates while falls were
only reported on 1.2\% of the death certificates. Conditions
listed on the left side are not associated with any other
condition or disease.}\label{Fig_graph_complet}
\end{center}
\end{figure}

\noindent Figure \ref{Fig_graph_complet} shows the final retained
model for the group 15-24, which is the intersection of the models
derived under methods \verb"SepLogit OR" and \verb"GaussCor".
Apart from the obvious association between depression and suicide,
the strongest (positive) associations identified were between
diabetes and other endocrinal diseases, colorectal cancer and
metastasis, septicemia and pneumonia, and between diseases of
arteries, arterioles and capillaries and cerebrovascular diseases.
The strong negative associations between transport accidents and
all other conditions, and between suicide and most other
conditions (except depression and other mental disorders) is also
worth noting. These latter associations correspond to well-known
sequences of causes leading to death and most of those present in
the figure have strong
biological plausibility. \vskip5pt 

\noindent In the analysis of the older group, we only applied
methods \verb"SepLogit" and \verb"GaussCor" (to save computational
time), which took 15241 and 4.8 seconds respectively. In this
case, we had to use
$\lambda^{\footnotesize{\mbox{min}}}=\lambda^{\footnotesize{\mbox{max}}}/50$
for method \verb"SepLogit". Moreover, when omitting the
computation of the un-shrunk estimates, the computational time
using method \verb"SepLogit" reduced to 1943 seconds. 600, 778 and
708 associations were detected by method \verb"SepLogit AND",
\verb"SepLogit OR" and \verb"GaussCor" respectively. Good
agreement was observed between the models (we had $\kappa=0.933$
for the comparison between \verb"GaussCor" and \verb"SepLogit OR"
and $\kappa=0.953$  for the comparison between \verb"GaussCor" and
\verb"SepLogit AND"), which tends to confirm that agreement
between the models returned by methods \verb"SepLogit" and
\verb"GaussCor" increases with the signal-to-noise ratio.

%

\section{Discussion}

\noindent In this paper we empirically compared several
approximate methods designed to search for associations among
binary variables. We observed that methods \verb"SepLogit" and
\verb"BMNPseudo" achieved similar performances in terms of
accuracy and F1-score, with a slight advantage to method
\verb"SepLogit". Moreover, the models selected by both methods are
very similar in most cases, as could be expected given the
similarity between them. In terms of computational time,
\verb"SepLogit" appeared to be overall faster than
\verb"BMNPseudo", but the two methods share the disadvantage of
being quite slow to compute, especially for low values of the
sparsity parameter. \vskip5pt

\noindent For the method \verb"BMNPseudo", we observed that using
half the pseudo-likelihood rather than the pseudo-likelihood
itself when computing the BIC enables us to select better models
in most cases. The multiplicative coefficient 1/2 might not be
optimal in all situation and some adaptive coefficient might be
derived from a theoretical study of the pseudo-likelihood.
Alternatively, cross-validation could be considered at a cost of
an increased computational time, which seems undesirable given the
aforementioned lack of speed of this method. Moreover, the
suitability of cross-validation for model selection remains
questionable \citep{Gaoetal}. \vskip5pt

\noindent In terms of accuracy, the method proposed by
\cite{Banerjee} was shown to be generally too conservative. We
then proposed a slight modification, referred to as
\verb"GaussCor", in which we remove the additive $1/3$ term on the
diagonal, and use the sample correlation matrix as a starting
point for the algorithm.  With these modifications,
\verb"GaussCor" combines good overall performances (comparable to
the performances achieved by \verb"SepLogit" and \verb"BMNPseudo")
and exceptional computational speed.  In particular,
\verb"GaussCor" was observed to be between 3 and 200 times faster
than the other methods on simulated data. This speed is
particularly desirable for handling truly high-dimensional
datasets since the concurrent methods (\verb"SepLogit" or
\verb"BMNPseudo") might be dramatically slow in such cases. To be
complete, we should mention that method \verb"SepLogit" could be
implemented using other sparse logistic regression algorithms that
might be faster than the \verb"glmnet" R package (see
\cite{Boyd_logreg,BBR,IRLSLARS} for instance). However, we think
that the comparison conducted here was fair since both R packages
\verb"glmnet" and \verb"glasso" rely on a coordinate descent
algorithm \citep{glmnet}. \vskip5pt

\noindent Interestingly, we also observed that the models selected
by methods \verb"GaussCor" and \verb"SepLogit" can be
significantly different, especially in the situation of low
signal-to-noise ratio. On our real example, we decided to retain
the intersection of the two selected models as the final model,
which led to conservative but competitive models on simulated
examples. However, other approaches can be considered and it would
be interesting to further study how these methods can be optimally
combined.\vskip5pt

\noindent Approximate methods that use either multiple logistic
regressions or the pseudo-likelihood have been shown to attain
performances similar to those reached using exact inference at a
lower computational cost \citep{Hofling}.
 Our results suggest here that using Gaussian approximates of the
Ising likelihood can ensure similar statistical performance at a
greatly improved speed. In the absence of a theoretical
justification for the good performances achieved by this method
however, we can only claim here that \verb"GaussCor" is a
candidate method that can be recommended in some cases; a
theoretical study might enable to have a better idea of what these
cases are.

\newpage

\appendix
\section*{Appendix: The retained classification of causes of death}

\begin{center}
\footnotesize \tablehead{\hline
{\bf Disease/Condition } & {\bf Label} & {\bf ICD-10 codes} & {\bf Count} \\
\hline}
\begin{supertabular}{p{5cm} c p{5cm} c}

Septicaemia & septicaemia &     A40-A41  &      25713 \\

tuberculosis &     tuberc &  A15-A19, B90  &       1720 \\

aids and HIV infection &       aids &     B20-B24  &       1094 \\

Other infectious disease & other\_infect &  A00-A14, A20-A39, A42-B19, B25-B89, B91-B99  &      14188 \\

Oral cancer &    c\_oral &     C00-C14  &       5076 \\

Oesophageal cancer &  c\_oesoph &         C15  &       4654 \\

Stomach cancer & c\_estomac &         C16  &       5642 \\

Colorectal cancer & c\_colorect &     C18-C21  &      19587 \\

Liver cancer &   c\_liver &         C22  &       8528 \\

Pancreas cancer & c\_pancreas &         C25  &       8615 \\

Larynx cancer &  c\_larynx &         C32  &       2062 \\

Lung cancer &    c\_lung &     C33-C34  &      30642 \\

Breast cancer &  c\_breast &         C50  &      14439 \\

Uterus cancer &  c\_uterus &     C53-C55  &       3708 \\

Prostate cancer & c\_prostate &         C61  &      13361 \\

Bladder cancer & c\_bladder &         C67  &       5946 \\

Hodgkin's disease and leukemia & c\_hodgkin &     C81-C96  &      15574 \\

Secondary malignant neoplasm & c\_metasta &     C77-C79  &      50314 \\

Other cancers &   c\_other &  C17, C23-C24, C26-C31, C35-C49, C51-C52, C56-C60, C62-C66, C68-C76, C80, C97-D49  &      72943 \\

Diseases of the blood & blood\_dis &     D50-D89  &      12955 \\

Diabetes &    diabetes &     E10-E14  &      32704 \\

Malnutrition &     malnut &     E40-E46  &      12713 \\

Other endocrinal disease & other\_endo &  E00-E09, E15-E39, E47-E90   &      27490 \\

Dementia &   dementia &     F01-F03  &      22966 \\

Mental disorders due to use of alcohol & alcohol\_ment &         F10  &      12634 \\

Mood disorders & depression &     F30-F39  &       8628 \\

Other mental disorders & other\_ment &  F00, F04-F09, F11-F29, F40-F99  &      15629 \\

Parkinson's disease &  parkinson &         G20  &       8598 \\

Alzheimer's disease &  Alzheimer &         G30  &      22568 \\

Other diseases of the nervous system,the eye and adnexa & other\_nervous &  G00-G19, G21-G29, G31-H95  &      27028 \\

Hypertensive diseases &  hypertens &     I10-I15  &      44117 \\

Ischaemic heart diseases & isc\_heart\_dis &     I20-I25  &      62071 \\

Pulmonary embolism,phlebitis and thrombophlebitis &   embolism &  I26, I80-I82  &      16697 \\

Cardiac arrhythmias & card\_arrhytm &     I47-I49  &      38020 \\

Heart failure & heart\_fail &         I50  &      73268 \\

Cerebrovascular diseases & cere\_vasc &     I60-I69  &      58161 \\

Diseases of arteries, arterioles and capillaries &   arteries &     I70-I79  &      25567 \\

Other diseases of the circulatory system & other\_circ &  I00-I09, I16-I19, I27-I46, I51-I59, I83-I99  &      80060 \\

Influenza (other than avian influenza) &  influenza &     J10-J11  &       1192 \\

Pneumonia &     pneumo &     J12-J18  &      38677 \\

Asthma and status asthmaticus &     asthma &     J45-J46  &       2886 \\

Other chronic lower respiratory diseases & other\_low\_resp &  J40-J44, J47  &      17739 \\

Lung diseases due to external agents & lung\_extern &     J60-J70  &      10137 \\

Other diseases of the respiratory system & other\_resp &  J00-J09, J19-J39, J48-J59, J71-J99  &      59123 \\

Peptic ulcer & pept\_ulcer &     K25-K28  &       1965 \\

Diseases of liver & liver\_dis &     K70-K77  &      20661 \\

Other diseases of the digestive system & other\_digest &  K00-K24, K29-K69, K78-K99  &      34507 \\

Diseases of the skin and subcutaneous tissue &  skin\_dis &     L00-L99  &      11056 \\

Diseases of the musculoskeletal system and connective tissue & musculoskeletal &     M00-M99  &       9644 \\

Diseases of the genitourinary system & genito\_urinary &     N00-N99  &      37683 \\

Pregnancy, childbirth and the puerperium & pregnancy &    O00-O99  &       67 \\

Certain conditions originating in the perinatal period & perinat\_cond &     P00-P96  &       2100 \\

Congenital malformations, deformations and chromosomal abnormalities & congenit\_malf &     Q00-Q99  &       2411 \\

Senility &   senility &         R54  &      23646 \\

Other symptoms and abnormal clinical findings, not elsewhere classified & other\_undef &  R00-R53, R55-R59  &     252979 \\

Transport accidents & transp\_acc &     V01-V99  &       5686 \\

Falls &      falls &     W00-W19  &       6217 \\

Intentional self-harm &    suicide &     X60-X84  &      10900 \\

Other external causes of morbidity and mortality & other\_extern &  W20-X59, X85-Y89  &      21813 \\

\hline
\end{supertabular}
\end{center}
\vskip 0.2in

\end{document}